\documentclass[acmsmall,screen=False]{acmart}

\AtBeginDocument{%
  \providecommand\BibTeX{{%
    \normalfont B\kern-0.5em{\scshape i\kern-0.25em b}\kern-0.8em\TeX}}}

\setcopyright{acmcopyright}
\copyrightyear{2022}
\acmYear{2022}
\acmDOI{10.1145/1122445.xxxxx}

\acmJournal{JACM}
\acmVolume{37}
\acmNumber{4}
\acmArticle{111}
\acmMonth{1}

\usepackage{microtype}
\usepackage{comment}
\usepackage{float}
\usepackage{xcolor}
\restylefloat{table}
\usepackage{amsmath}
\usepackage[utf8]{inputenc}
\usepackage{graphicx}

\usepackage{adjustbox}
\usepackage{multirow}
\usepackage{xhfill}
\usepackage{arydshln}
\usepackage{soul}

\makeatletter
\def\thickhline{%
  \noalign{\ifnum0=`}\fi\hrule \@height \thickarrayrulewidth \futurelet
   \reserved@a\@xthickhline}
\def\@xthickhline{\ifx\reserved@a\thickhline
               \vskip\doublerulesep
               \vskip-\thickarrayrulewidth
             \fi
      \ifnum0=`{\fi}}
\makeatother

\newlength{\thickarrayrulewidth}
\newcommand{\hlt}[1]{\textcolor{black}{#1}}

\setcopyright{acmcopyright}
\acmJournal{CSUR}
\acmYear{2022} \acmVolume{1} \acmNumber{1} \acmArticle{1} \acmMonth{1} \acmPrice{15.00}\acmDOI{10.1145/3545176}

\begin{document}

\title{An Empirical Survey on Long Document Summarization: Datasets, Models and Metrics}


\author{Huan Yee Koh}
\email{huan.koh@monash.edu}
\affiliation{%
  \institution{Monash University}
  \country{Australia}
}
\author{Jiaxin Ju}
\email{jjuu0002@student.monash.edu}
\affiliation{%
  \institution{Monash University}
  \country{Australia}
}

\author{Ming Liu}
\authornote{Corresponding Authors: Ming Liu and Shirui Pan.}
\affiliation{%
  \institution{Deakin University}
    \country{Australia}
}
\email{m.liu@deakin.edu.au}

\author{Shirui Pan}
\authornotemark[1]
\authornote{This work is done while Shirui Pan is with Monash University. From August 2022, he is with the School of Information and Communication Technology, Griffith University, Southport, QLD 4222, Australia.}

\affiliation{%
  \institution{Monash University}
  \country{Australia}
}
\email{shiruipan@ieee.org}



\begin{abstract}
Long documents such as academic articles and business reports have been the standard format to detail out important issues and complicated subjects that require extra attention. An automatic summarization system that can effectively condense long documents into short and concise texts to encapsulate the most important information would thus be significant in aiding the reader's comprehension. Recently, with the advent of neural architectures, significant research efforts have been made to advance automatic text summarization systems, and numerous studies on the challenges of extending these systems to the long document domain have emerged. In this survey, we provide a comprehensive overview of the research on long document summarization and a systematic evaluation across the three principal components of its research setting: benchmark datasets, summarization models, and evaluation metrics. For each component, we organize the literature within the context of long document summarization and conduct an empirical analysis to broaden the perspective on current research progress. The empirical analysis includes a study on the intrinsic characteristics of benchmark datasets, a multi-dimensional analysis of summarization models, and  a review of the summarization evaluation metrics. Based on the overall findings, we conclude by proposing possible directions for future exploration in this rapidly growing field.

\end{abstract}
\maketitle
\section{Introduction}
Summarization of textual information is an exacting task for humans and the rate of information growth in the era of big data has made summarizing most information manually to be impractical and impossible. This phenomenon is exacerbated when it comes to long form textual documents as the knowledge and human labour effort required to process and summarize it increases exponentially with the length of documents. Inevitably, a significant amount of invaluable information and knowledge have gone unnoticed, presenting an important bottleneck in the progress of social and economic development. In response, there has been a strong demand for exhaustive research in the field of automatic long document summarization \cite{cohan2018discourse,sharma2019bigpatent,beltagy2020longformer,dong2021discourse,manakul2021longspan}. 

Automatic text summarization involves a process of shortening a source text efficiently while keeping the main idea intact, which aids in reducing the amount of time required to process information, helps with faster search for information, and makes learning one topic easier \cite{liu2019automatic,liu2018generating}. While the potentiality of developing an effective automatic text summarization system has attracted significant interest and attention from the research community, automatic text summarization remains a challenging task and is not ready for wide practical use in day-to-day lives, particularly when it comes to summarizing long documents \cite{cao2018faithful,kryscinski2019neural,maynez2020faithfulness,kryscinski-etal-2020-evaluating}. Intuitively, long document summarization is harder than short document summarization due to the significant difference in the amount of lexical tokens and breadth of content between short and long documents. As the length increases, the content that would be considered important will also increase, resulting in a more challenging task for an automatic summarization model to capture all salient information in the limited output length \cite{gidiotis2020divide}. Further, short documents are often generic text such as news articles \cite{sandhaus2008new,nallapati2016abstractive,narayan2018don,grusky2018newsroom}, while long documents are commonly domain-specific articles such as scientific papers that contain more complex formulas and terminologies \cite{cohan2018discourse,kornilova2019billsum,huang2021efficient}. Together with other reasons that will be explored in this survey, long document summarization poses a significantly more challenging task than short document summarization. 

In general, automatic text summarization can be conceptualized as having three approaches: extractive, abstractive, and hybrid approach \cite{kryscinski-etal-2020-evaluating}. The extractive approach directly copies salient sentences from the source document and combine them as the output \cite{gong2001generic,cheng2016neural}, whereas the abstractive approach imitates human that comprehends a source document and writes a summary output based on the salient concepts of the source document \cite{rush2015neural,see2017get}. The hybrid approach attempts to combine the best of both approaches by rewriting a summary based on a subset of salient content extracted from the source document \cite{hsu2018unified,liu2018generating,gehrmann2018bottom}. Each approach has its advantages and limitations that may suit certain summarization tasks better. For example, extractive summarization may be sufficient in summarizing certain news articles \cite{cheng2016neural,zhang2018abstractiveness} but inadequate to summarize a long dialogue where salient content are sparsely distributed \cite{zhang2021exploratory}. This is because while the extractive summarization approach is always factually consistent with the source document, it does not modify the original text and thus lacks the ability to generate fluent and concise summary \cite{xu2019discourse}.

Historically, to measure the performance of different summarization architectures, ROUGE score \cite{lin2004rouge} has been the modus operandi for researchers in the summarization research field to compare and study the quality of different candidate summaries. The core idea of ROUGE score is to measure the lexical overlaps such as words and phrases between candidate summary and ground truth summary. While it is efficient, recent findings have shown that ROUGE score does not correlate well with how humans assess the quality of a candidate summary \cite{kryscinski2019neural,bhandari2020re,chaganty-etal-2018-price,hashimoto-etal-2019-unifying}. As a result, there is a significant amount of effort in improving the way we measure the quality of candidate summaries and performance of summarization architectures \cite{kryscinski-etal-2020-evaluating,maynez2020faithfulness,zhang2019bertscore,mao2020facet,yuan2021can}. Unfortunately, these efforts have entirely been focusing on the short document domains and the progress in measuring the quality of long document summarization approach has been lacking \cite{pagnoni-etal-2021-understanding,graham2015re,huang2020have,bhandari2020re,peyrard2019studying}. 

Nevertheless, there has been a considerable amount of advancement made in the long document summarization research field and the area lacks a comprehensive survey \cite{gambhir2017recent,boorugu2020survey,el2021automatic,shi2021neural}. Our paper fills this gap by providing a comprehensive overview of the research on long document summarization and a systematic evaluation across the three principal components of its research setting: benchmark datasets, summarization models, and evaluation metrics.
\\
The contribution of our paper is as follows: 

\textit{Comprehensive Review.} A comprehensive survey of the long document summarization research literature. 

\textit{Full-view of summarization research.} Text summarization literature mainly explores the three key aspects of research setting: developing advanced models, releasing new datasets, and proposing alternative evaluation metrics. We empirically provide a detailed review of all three key components within the context of long document summarization.

\textit{Empirical Studies and Thorough Analysis.} To ensure wide coverage of emerging trends, we empirically analyze each component of the long document summarization research setting through fine-grained human analysis and ad-hoc experiments.

\textit{Future Direction.} We discuss the current progress of long document summarization, analyze the limitation of existing methods, and suggest promising future research directions in terms of model designs, quality and diversity of datasets, the practicality of evaluation metrics and, finally, the feasibility of implementing summarization techniques to real-life applications.

The survey is organized as follows: firstly, an overview of the fundamentals of long document summarization in section 2. Secondly, a detailed study of ten summarization benchmark datasets is in section 3. A comprehensive survey on summarization models that are designed specifically or have to ability to summarize long documents in section 4. Then, in section 5, we analyze the performances of models that are representative of the different types of architectures commonly used by researchers through ad-hoc experiments. In section 6, we summarize the advancement in evaluation metrics and their applicability in the long document summarization domain. Section 7 goes into the applications of long document summarization models and Section 8 discusses promising future research direction in this field. Finally, section 9 concludes this survey.
\section{Fundamentals of Long Document Summarization}
To make clear the distinction between short and long documents, we conceptualize the summarization task problem from three different fundamental aspects: 1) length of document, 2) breath of content, and 3) degree of coherence. 
\subsection{Length of Document}
Documents are commonly classified as "long" because the number of lexical tokens in the source document is enormous and it requires a considerable amount of time for an average human to consume the full text. While this definition makes intuitive sense, in the context of machine learning, a document is considered long when current state-of-the-art models for a normal document cannot be implemented similarly in an effective manner due to hardware and model limitations. For example, previous research \cite{celikyilmaz2018deep} considers CNN/DM and NYT benchmark datasets in the news domain as long documents when in the present research context they are now considered to be short document datasets. Currently, a benchmark dataset with an average source document length that exceeds 3,000 lexical tokens could be well-considered as "long documents" \cite{zaheer2020big,manakul2021longspan} due to the fact that most existing state-of-the-art summarization systems (e.g., pre-trained models) are limited to 512 to 1024 lexical tokens only \cite{devlin2019bert,zhang2020pegasus}. These limitations cannot be easily solved without novel techniques that help in assisting current architectures to reason over a long range of textual inputs \cite{zaheer2020big, manakul2021longspan,meng2021bringing}. Accordingly, this survey adopts a similar definition where a document is only considered as long if current state-of-the-art systems used in the short document cannot be extended and applied to a document with significantly longer text. Despite the potentially confusing definition, this enduring definition ensures that the model architectures implemented by researchers require novel techniques to overcome hardware limitations rather than just a mere replica of previous works.    
\subsection{Breadth of Content}
On average, informative content that is non-redundant will increase together with the length of a document. However, despite the fact that reference summary length often increases together with the source document length, the length of a summary is usually constrained by what an average user considered as reasonable \cite{cohan2018discourse,sharma2019bigpatent}. Thus, while it may be sufficient for summaries of a reasonable length to cover the most or even all of the informative aspects for short documents, this is not necessarily true for summaries of long documents. In section 3, we empirically show that the relative length of summary against the source document becomes exponentially shorter as the source document length increases. Due to this elevated constraint, the ground truth summary of a long document will inevitably lose information that is not key to the central narrative of the original author or summary writer \cite{gidiotis2020divide}. Furthermore, recent work \cite{kryscinski2019neural} has also identified that human users could not agree on what should be considered important for a given document in the short document news domain due to the heterogeneity of user preferences and expectations. This issue is exacerbated when it comes to long document summarization as (a) the relative length of summary against the source document is shorter and (b) the chance of users having different preferences and expectations would increase as the breadth of content increases, making the long document summarization task significantly harder than short document. 
\subsection{Degree of Coherence}
As compared to short documents, long documents are often structured into sections for the ease of user comprehension \cite{cohan2018discourse,kornilova2019billsum}. The content within each section also differ to a certain extent despite revolving around a key narrative of the long documents. This makes the long document summarization task more burdensome as summarization models cannot concatenate salient texts from different sections without considering its impact on the fluency, redundancy, and semantic coherence of the final summary outputs. \\\\Based on the fundamental aspects, the rest of this paper provides an empirical survey on long document summarization, covering the benchmark datasets, summarization models, and metrics. 
\section{Datasets}
Publicly available benchmark datasets have been introduced to evaluate the performance of summarization models. Nonetheless, the benchmark datasets have different intrinsic characteristics that have been found to be crucial in the understanding of model performances \cite{maynez2020faithfulness,tejaswin2021well}, summarization approach suitability (i.e., extractive or abstractive approach) \cite{zhang2018abstractiveness,sharma2019bigpatent} and evaluation metrics effectiveness \cite{gabriel-etal-2021-go,pagnoni-etal-2021-understanding}. Hence, only through a comprehensive understanding of the benchmark datasets, one can assess the underlying performance and applicability of a summarization model in the real-world settings \cite{kryscinski2019neural}. Further, insights drawn from benchmark datasets have led to the introduction of state-of-the-art models across a wide range of natural language processing (NLP) tasks  \cite{chen2016thorough,yatskar2019qualitative}, including the text summarization task \cite{gidiotis2020divide, dong2021discourse,manakul2021longspan}. In response, intrinsic dataset evaluation through large-scale automatic evaluation \cite{bommasani2020intrinsic} or more fine-grained human evaluation at a smaller scale \cite{tejaswin2021well} has also been performed to enhance the understanding of various benchmarks. Nevertheless, none of the aforementioned works performed a large-scale automatic evaluation analysis nor a thorough human evaluation of benchmark datasets in the long document text summarization domain. To address this gap, this section explores the basic statistics and intrinsic characteristics of popular benchmarks in short and long document domain through the usage of large-scale automatic evaluation metrics and performs fine-grained human analysis on the arXiv benchmark to encourage a better appreciation of the most widely used long document summarization dataset \cite{zaheer2020big,huang2021efficient,manakul2021longspan}. 

\subsection{Corpora}
\paragraph{Short document} 
Short-document datasets studied in this survey are CNN-DM, NWS, XSUM, Reddit-TIFU, and WikiHow. The first three news datasets are chosen due to their popularity while Reddit-TIFU and WikiHow are studied to ensure short documents from other domains are also included. The document-summary pairs from CNN-DM, NWS, and XSUM are typical of that in the news domain, where the source document represents news article while the summary represents either human-curated summary \cite{grusky2018newsroom,narayan2018don} or summary created by concatenating bullet-point sentences in the original source document \cite{nallapati2016abstractive}. On the other hand, Reddit-TIFU is a dataset collected from the subreddit r/TIFU \cite{kim2019abstractive}, while the WikiHow benchmark is created using the first sentence of each WikiHow web page's paragraph as the summary and the rest as source text \cite{Koupaee2018WikiHowAL}.

\paragraph{Long document} 
For long document summarization research, arXiv, PubMed, BIGPATENT, BillSum, and GovReport have been used in prior research to test and compare novel long document summarization models. arXiv and PubMed \cite{cohan2018discourse} are scientific long document summarization datasets collected from arXiv.org and PubMed.com scientific repository. Both datasets represent the earliest work on large-scale long document summarization datasets. BIGPATENT \cite{sharma2019bigpatent} is an enormous dataset with over 1.3 million document-summary records of U.S. patent documents along with human written abstractive summary. BillSum \cite{kornilova2019billsum} is a dataset on summarizing Congressional and California state bills where the content structures and stylistic features of writing are considerably different from documents in other domains. GovReport \cite{huang2021efficient}, assembled from reports published by U.S. Government Accountability Office, is markedly longer than the other long document datasets. Other long document benchmark datasets that are worth mentioning but are no longer widely used due to the limited amount of document-summary pairs are CL-SciSumm and SciSummNet  \cite{jaidka-etal-2016-overview,yasunaga2019scisummnet}. Some other benchmark datasets that are released more recently in the podcast \cite{clifton-etal-2020-100000} and dialogue domains \cite{janin2003icsi,carletta2005ami,rameshkumar-bailey-2020-storytelling,zhu2021mediasum} may also be classified as long document summarization benchmark \cite{manakul2021longspan} but are not explored in this survey as dialogue summarization has been recognized as another sub-domain due to its distinctive features as compared to other document types.

\subsection{Data Metrics} Given a document, $D$, and a corresponding reference summary, $S$, each document will have a sequence of tokens $D_{token} = \{ t_1, t_2, ..., t_n \} $ and each summary will also have a sequence of tokens $ S_{token} = \{ t^*_1, t^*_2, ..., t^*_m \} $. Similarly, each document and summary have $l$ and $o$ sentences as represented by $D_{sent} = \{ s_1, s_2,...,s_{l} \}$ and $S_{sent} = \{ s^*_1, s^*_2,...,s^*_o \}$ respectively. Length of document and summary measured in number of tokens are represented as $|D|$ and $|S|$ while length measured in number of sentences are represented as $||D||$ and $||S||$. Extending on the works in the short document summarization domain \cite{bommasani2020intrinsic, grusky2018newsroom}, the following discusses each of the five metrics used to evaluate the benchmark datasets shown in Table \ref{tab: datasets}: compression ratio, extractive coverage, extractive density, redundancy and uniformity. 

\textit{Compression Ratio} measures the ratio of a source document length against its reference summary length. A higher compression ratio indicates larger information loss in the original document after being summarized. Compression ratios are measured based on tokens and sentences:

$$COMPRESSION_{token} = \frac{|D|}{|S|} \quad\text{  and  }\quad COMPRESSION_{sent} = \frac{||D||}{||S||}$$

\textit{Extractive Coverage and Extractive Density} are introduced by \citet{grusky2018newsroom} based on the notion of matching fragments. Fragments are obtained by greedily matching the longest shared token sequence where $\mathcal{F}(D,S)$ reflects a set of fragments with each fragment having a length represented by $|f|$. Extractive coverage calculates the percentage of tokens in summary that is a derivation of the original source text, whereas, extractive density relates to the average squared length of the extractive fragments in the summary. The former indicates the need for a model to coin novel tokens that are not in the original source text while the latter measures whether a model can match the ground truth summary merely by extracting from the original source text without rearranging or paraphrasing text. 
$$COVERAGE(D,S) =  \frac{1}{|S|} \sum_{f \in \mathcal{F}(D,S)} |f| $$
$$DENSITY(D,S) = \frac{1}{|S|} \sum_{f \in \mathcal{F}(D,S)} |f|^2 $$

\textit{Redundancy} relates to the redundancy of ground truth summary by measuring the average ROUGE-L F1-score of all distinct pairs of summary sentences \cite{bommasani2020intrinsic}. As ROUGE-L measures the longest common sub-sequence overlap between two texts \cite{lin2004rouge}, a higher redundancy score would suggest that a candidate summary is more redundant as the sentence pairs in the ground truth summary contain more similar content in each sentence pair. For each summary consisting of m sentences,  $\mathbb{S}$, we have a set of distinct pairs of sentences, $\mathbb{S} \times \mathbb{S} $, where the redundancy score is calculated as:  
$$ REDUNDANCY(S) = \underset{(x_i,x_j) \in \mathbb{S}\times\mathbb{S},\, x_i \neq x_j}{average}  ROUGE(x_i,x_j) $$

\textit{Uniformity} measures whether content that are considered important by the reference summary are uniformly scattered across the entire source document. A higher score indicates that important content are scattered across the entire document with no obvious layout bias to take advantage of. This is calculated based on the normalized entropy of the decile positions of salient unigrams in the source text, where salient unigrams are the top 20 keywords extracted\footnote{We use NLTK-RAKE for keywords extraction.}, excluding stopwords, from the reference summary.  
$$UNF(unigram_{pos})= H_{norm}(unigram_{pos}) $$

\subsection{Intrinsic Characteristics of Datasets}
\subsubsection{Short vs Long Document Benchmark Dataset}\,\\
Based on Table \ref{tab: datasets} below, the following discusses the \textbf{findings} of intrinsic characteristics of long document benchmark datasets in comparison to short document benchmark datasets. 

\textbf{Finding 1. Length of Long Documents:} A basic yet important finding is that, except for BillSum, all the other long document datasets have an average source document length of at least 3,000 tokens. In contrast, the longest short document dataset, CNN-DM, has an average document length of 774 tokens. This indicates that a vanilla pre-trained Transformer-based models \cite{raffel2020exploring,lewis2020bart,zhang2020pegasus} 
which commonly have an input length limit of 1,024 tokens would need to truncate at least half of the source document in the long document benchmark datasets. Thus, if pre-trained models that have proven to work well under short document settings are implemented without any long document adaptations in their architectural settings and mechanisms, they are unlikely to generate high-quality summaries for long documents \cite{maynez2020faithfulness,he2020ctrlsum,rothe-etal-2021-thorough}.

\begin{table*}[!ht]
\centering
\resizebox{\textwidth}{!}{%
\small
\begin{tabular}{c|ccccc|ccccc|c}
\thickhline
\multicolumn{1}{l|}{} & \multicolumn{5}{c|}{\textbf{Short Document Datasets}} & \multicolumn{5}{c|}{\textbf{Long Document Datasets}} & {\textbf{Long vs. Short}}\\ [0.5ex]
\multicolumn{1}{l|}{} & CNN-DM & NWS & XSum & WikiHow & Reddit & ArXiv & PubMed & BigPatent & BillSum & GovReport & Avg. Ratio \\ [0.3ex]
\hline
\textbf{\# doc-summ.} & 278K & 955K & 203K & 231K & 120K & 215K & 133K & 1.34M & 21.3K & 19.5K & - \\
\textbf{summ tokens} & 55 & 31 & 24 & 70 & 23 & 242 & 208 & 117 & 243 & 607 & 6.9x \\
\textbf{doc tokens} & 774 & 767 & 438 & 501 & 444 & 6446 & 3143 & 3573 & 1686 & 9409 & 8.3x \\
\textbf{summ sents} & 3.8 & 1.5 & 1 & 5.3 & 1.4 & 6.3 & 7.1 & 3.6 & 7.1 & 21.4 & 3.7x \\
\textbf{doc sents} & 29 & 31 & 19 & 27 & 22 & 251 & 102 & 143 & 42 & 300 & 6.5x \\ [0.3ex]
\hline
\textbf{Compression\textsubscript{token}} & 14.8 & 31.7  & 19.7 & 7.2 & 18.4 & 41.2 & 16.6 & 36.3 & 12.2 & 18.7 &  1.4x \\
\textbf{Compression\textsubscript{sent}} & 8.3 & 22.4 & 18.9 & 3.3 & 14.5 & 44.3 & 15.6 & 58.7 & 9.7 & 18.1 & 2.2x \\
\textbf{Coverage} & 0.890 & 0.855 & 0.675 & 0.610 & 0.728 & 0.920 & 0.893 & 0.861 & 0.913 & 0.942 & 1.2x \\
\textbf{Density} & 3.6 & 9.8 & 1.1 & 1.1 & 1.4 &  3.7 & 5.6 & 2.1 & 6.6 & 7.7 & 1.5x \\
\textbf{Redundancy} & 0.157 & 0.088 & - & 0.324 & 0.078 & 0.144 & 0.146 & 0.223 & 0.163 & 0.124 & 1.0x \\
\textbf{Uniformity} & 0.856 & 0.781 & 0.841 & 0.813 & 0.777 & 0.894  & 0.896 & 0.922  &  0.903 & 0.932 & 1.2x \\
\thickhline
\end{tabular}%
}
\caption{Comparison of Short and Long Document Summarization Datasets. Intrinsic characteristics are computed based on the average result of test samples. Average Ratios are computed based on the average long over short document statistics.}
\label{tab: datasets}
\end{table*}

\textbf{Finding 2. High Compression Ratio and its Implications:} On average, the token-level and sentence-level compression ratio of the long document summarization datasets is greater than the short document datasets by 1.4 and 2.2 times respectively. For long documents, this suggests that either a) there is a greater information loss in the summaries, b) the salient content is more sparsely distributed across the source documents, and/or c) the source document contains significantly more redundant information. As the high compression ratio of long document benchmark is more likely to be the results of the two former factors, this increases the relative difficulty of the long document summarization task as a model would have to clearly identify the key narrative from the source while excluding the content that are \textit{expected} to be less important by the summary readers. Moreover, if there is a greater information loss in the summary of a long document, the generated summary will inevitably miss an even greater amount of information that is considered important by some readers, diminishing the effectiveness of a generalized summarization approach to satisfy the needs of summary readers. This finding supports the efforts in controllable summarization, where the final generated summaries will be based on the reader's needs and expectations \cite{he2020ctrlsum,wu-etal-2021-controllable}.

\textbf{Finding 3. Abstractiveness and Diversity of Datasets:} With the exception of BIGPATENT, all long document datasets have greater coverage and density values than the short document datasets. This is likely due to the genres of benchmark datasets where long documents are often related to domain-specific articles such as scientific papers that contain more complex formulas and terminologies. Nonetheless, this indicates that a model that merely extracts lexical fragments from the original source text of a long document can still generate a summary that more closely resembles the reference summary.  As abstractive summarization models have recently been found to contain factual inconsistencies in up to 30\% of the summary outputs in the short document domain \cite{cao2018faithful,kryscinski-etal-2020-evaluating} while extractive summarization model will faithfully preserve the original content, this finding is encouraging for the development of long document extractive models in the real-world production level settings. Finally, as the abstractiveness of datasets have been found to greatly affect the summarization strategies of a supervised model \cite{zhang2018abstractiveness,xu2020understanding,wilber-etal-2021-point}, efforts to introduce benchmark datasets with greater abstractiveness (low extractive coverage and density value) should be encouraged to improve the diversity of long document benchmark datasets. 

\textbf{Finding 4. Lesser Layout Bias in Long Document:} \citet{kryscinski2019neural} found substantial layout bias in the source text where nearly 60\% of important sentences are contained 
in the first 30\% of the source articles and argued that such layout bias does not apply to the other domains. Our findings on the uniformity of salient content in Table \ref{tab: datasets} validates their arguments where the salient content of long documents are scattered across the entire source text more uniformly than the short documents. This suggests that unlike practices in the short document summarization domain where models are often benefited by taking advantage of layout biases \cite{see2017get,gehrmann2018bottom,paulus2018deep}, long document models that implement a truncation strategy to process only a small subset of the leading content of the long documents will likely suffer from significant performance degradation. 

\begin{figure} [!ht]
    \centering
       \includegraphics[width=1\textwidth]{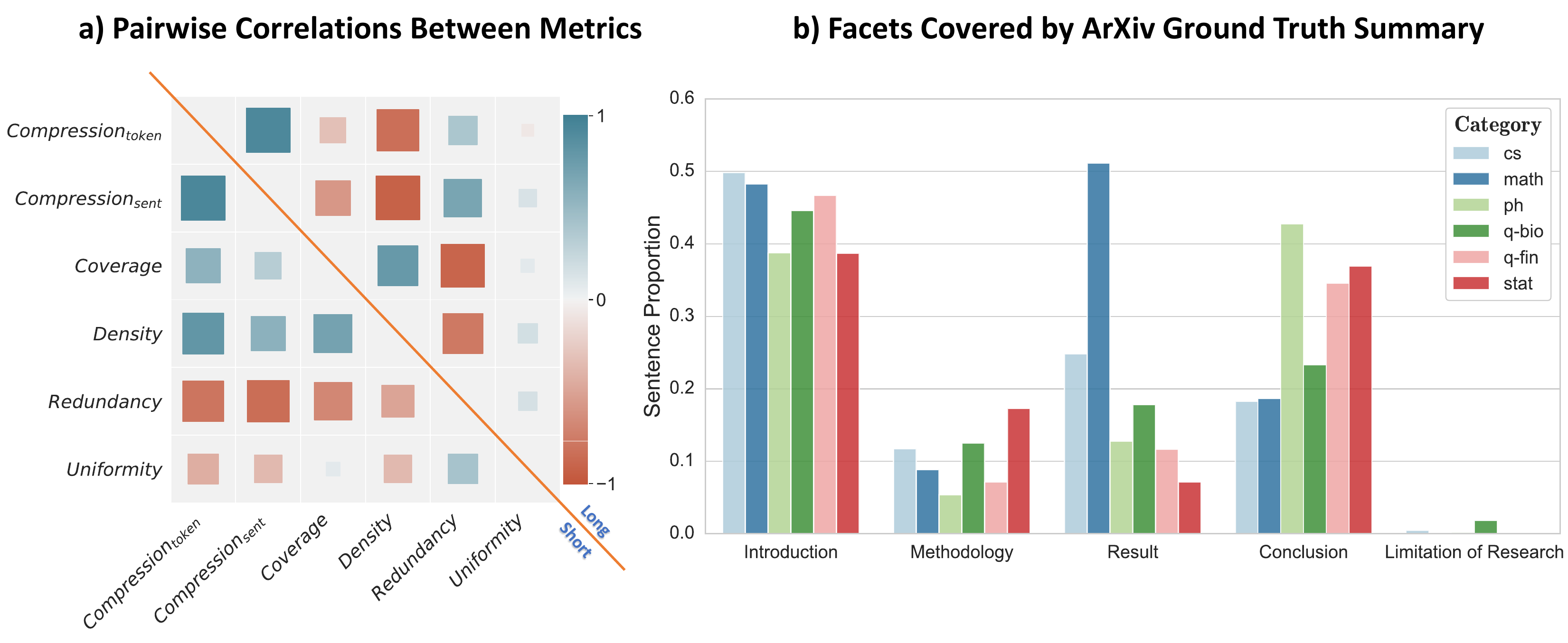}\\
       \caption{Pairwise Correlations between Metrics (figure 1a) and Facets Covered by ArXiv Ground Truth Summary (figure 1b). For figure 1a, the upper diagonal reflects Pearson correlation coefficient for long document benchmarks while the lower diagonal reflects values for short document benchmarks. Figure 1b's barplot is constructed based on human annotated testset data as described in section 3.5.}
       \label{fig: pairwise_corr}
\end{figure}

\textbf{Finding 5. Relationship between Intrinsic Characteristics:}
Other than the intrinsic characteristic measured in Table \ref{tab: datasets}, the statistical relationship between these metrics could yield insights regarding the underlying properties of a benchmark dataset. More importantly, whether the relationship between these metrics differs significantly under short and long document summarization settings should also be of great interest to practitioners. To quantify this, we report the pairwise correlations between each metric pair for both short document (lower diagonal) and long document (upper diagonal) benchmark datasets in figure \ref{fig: pairwise_corr}a. The values reported are calculated using the Pearson correlation coefficient, $\rho$. As represented by darker blue color in figure \ref{fig: pairwise_corr}a, $\rho = 1$ reflects a perfectly positive correlation between the metric pair and $\rho = -1$ when it is perfectly negative (shown in darker red color).

For positive controls, we see a strong positive relationship between the two compression ratios and the two extractive metrics (coverage and density) under short and long document settings. We also see a lack of statistical correlation when uniformity is measured against other metrics as uniformity relates more to the genres of documents rather than the other characteristics. We further observe redundancy to be inversely related to coverage and density, where a more abstractive reference summary often contains more redundant information. This finding is consistent with a human evaluation study by \citet{kryscinski-etal-2020-evaluating} where writers are found to be more verbose and write summary content that do not add information when they are writing unconstrained, abstractive summaries. Intriguingly, we see a weakly positive correlation between the extractive metrics and the compression metrics under the short document setting but a strongly negative correlation under the long document setting. It is hypothesize that when authors have to write a concise summary, they are forced to paraphrase the original content more to ensure that the summary can cover the salient content within the constrained summary length.

\subsubsection{Comparison between Long Document Dataset Benchmarks}\,\\
Looking at the intrinsic characteristics between long document benchmark datasets, arXiv and BIGPATENT have significantly higher compression ratios but lower extractive density values than the others, indicating that two of these datasets require a summarization model to generate a significantly shorter summary that is not written in the same way as the source text. As discussed above, this is likely because for a summary to cover more content within a constrained summary length, one has to paraphrase the original content more. This is also evidenced by the diverging values between extractive coverage and density metric for arXiv benchmark dataset that suggest summaries of arXiv scientific papers have high matching tokens and terminologies with the source document (high coverage) but low matching phrases (low density). Overall, the BIGPATENT dataset is the most suitable benchmark for long document supervised abstractive summarization due to its low coverage and density, substantial training sample pairs to serve as supervisory signals, and high uniformity in salient content. However, only a handful of fully-supervised abstractive long document summarization works \cite{pilault2020extractive,zaheer2020big,zhang2020pegasus} evaluated their models on BIGPATENT, limiting the visibility of current progress on long document summarizers in general applications. This is despite the fact that BIGPATENT was introduced not long after arXiv and PubMed. Encouragingly, with the recent introduction of long document datasets in domains other than scientific papers including financial reports \cite{loukas2021edgarcorpus} and books \cite{kryscinski2021booksum}, the research progress of long document summarization models towards general application should become clearer in the near future.

\subsection{Fine-grained Analysis on ArXiv}
To perform fine-grained human analysis on the arXiv benchmark, this survey implements a stratified random sampling strategy based on the 6 different categories of scientific domains contained in the arXiv.org scientific repository: physics (ph), computer-science (cs), mathematics (math), quantitative-biology (q-bio), quantitative-finance (q-fin) and statistics (stat). In total, we obtain over 700 annotated ground truth summaries with physics having the most samples (369) followed by computer science (140). Based on fine-grained human analysis of 743 ground truth summaries in the arXiv test set, this subsection reports the disturbing data quality results and studies the degree of diversity in formatting style of reference summary. 
\paragraph{Noise in arXiv benchmark dataset:}
With the advent of data-hungry neural architectures, there has been an enormous demand for benchmark datasets with document-summary pairs that are at least in the tens of thousands created through heuristic means such as scraping it directly from the web. As a result, depending on the means of extracting these datasets, the quality of benchmark datasets may vary significantly from one another. To this end, \citet{kryscinski2019neural} have quantified the percentage of samples with noise for CNN-DM and Newsroom from the short document summarization datasets to be 4.19\% and 3.17\% using simple heuristic methods. The noises found in the datasets through heuristic means can only suggest a lower bound of what the true amount of noises are as heuristic approaches can only detect obvious structural flaws in the samples. This suggests that the true underlying noises are extremely widespread and often understated. Glaringly, in our experiment, the problem of noisy data affects more than 60\% of the annotated ground truth summaries in the randomly sampled arXiv test set. This is greater than 54\% detected in XSUM dataset \cite{tejaswin2021well}. While many of the errors and noises are minor, more than 15\% of the reference summaries have significant errors where at least half of the summary contains errors, rendering the summaries to be unreadable. Reassuringly, the rest of the test sets with identified noises are not overly significant and often only affect one or two sentences in a benchmark dataset with an average of 10 sentences in the reference summary. To further understand why the noises and errors occurred, we trace the original data based on the arXiv id provided by the benchmark datasets. It was found that many errors occurred such as missing content or sentence breaking after a newline could be due to large-scale scraping of the original data using pandoc \cite{cohan2018discourse}. As neural summarization models may overfit to these problematic noises and contribute to less interpretable benchmarking results, we release the annotated data to allow a better understanding of the common noises and to encourage quality improvement in future benchmark datasets\footnote{The annotated dataset are released: https://github.com/huankoh/long-doc-summarization}.

\paragraph{\hlt{Content Coverage of Reference Summary:}} 
Other than analyzing the noises in the arXiv benchmark dataset, we also explore 
to what extent the ground truth summary covers various sections or facets of the source article. Figure \ref{fig: pairwise_corr}b shows the average distribution of sections covered by ground truth summaries for different domains. The distribution is plotted based on the assumption that each sentence in the reference summary covers a single facet of the original article and human annotators are asked to identify the section covered by the summary sentences. The facets studied are Introduction, Methodology, Result, Conclusion and Limitation of Research. Interestingly, while the reference summaries in all domains have covered introduction and methodology sections with similar emphasis, we see a negative correlation between contribution and conclusion (i.e., papers that emphasize contributions will write less on conclusions, and vice-versa). Notably, we see scientific papers in the mathematics domain emphasize more on the contribution while papers in the physics domain emphasize more on conclusions. These results make intuitive sense as findings in mathematics often do not lead to a strong substantive conclusion. The trade-off between various sections also illustrates the inevitable information loss when summarizing a long document as the summary can only describe certain aspects of the source document but not all. 

\paragraph{\hlt{Style of Writing and External Knowledge:}}
Importantly, except for quantitative-biology, all scientific papers do not discuss the limitations of their research. This is consistent with common practices of writing abstracts to attract readers in reading the original paper by emphasizing on the result findings and contributions of the authors. Nevertheless, most researchers would find a discussion on the limitations of research works to be informative and significant. Whether the abstract itself represents the best possible summary for a summarization architecture to imitate from and learn how to appropriately summarize all the salient content including the limitation discussed in the original paper remains an important question to be answered. Recent progress on summarization approaches that generate user-specific summaries based on the need of readers are also important directions towards general applicability of summarization models in commercial settings \cite{he2020ctrlsum,wu-etal-2021-controllable}. Lastly, to summarize the limitations of a paper often requires more external knowledge outside of the content related to the source document and whether current summarization models are able to infer such knowledge from the benchmark dataset is an interesting study left for future works.   
\section{Models}
\subsection{Overview}
The following describes the differences between the extractive, abstractive and hybrid summarization approaches and the general taxonomy of a summarization system.

\begin{figure} [h!]
    \centering
       \includegraphics[width=0.9\textwidth]{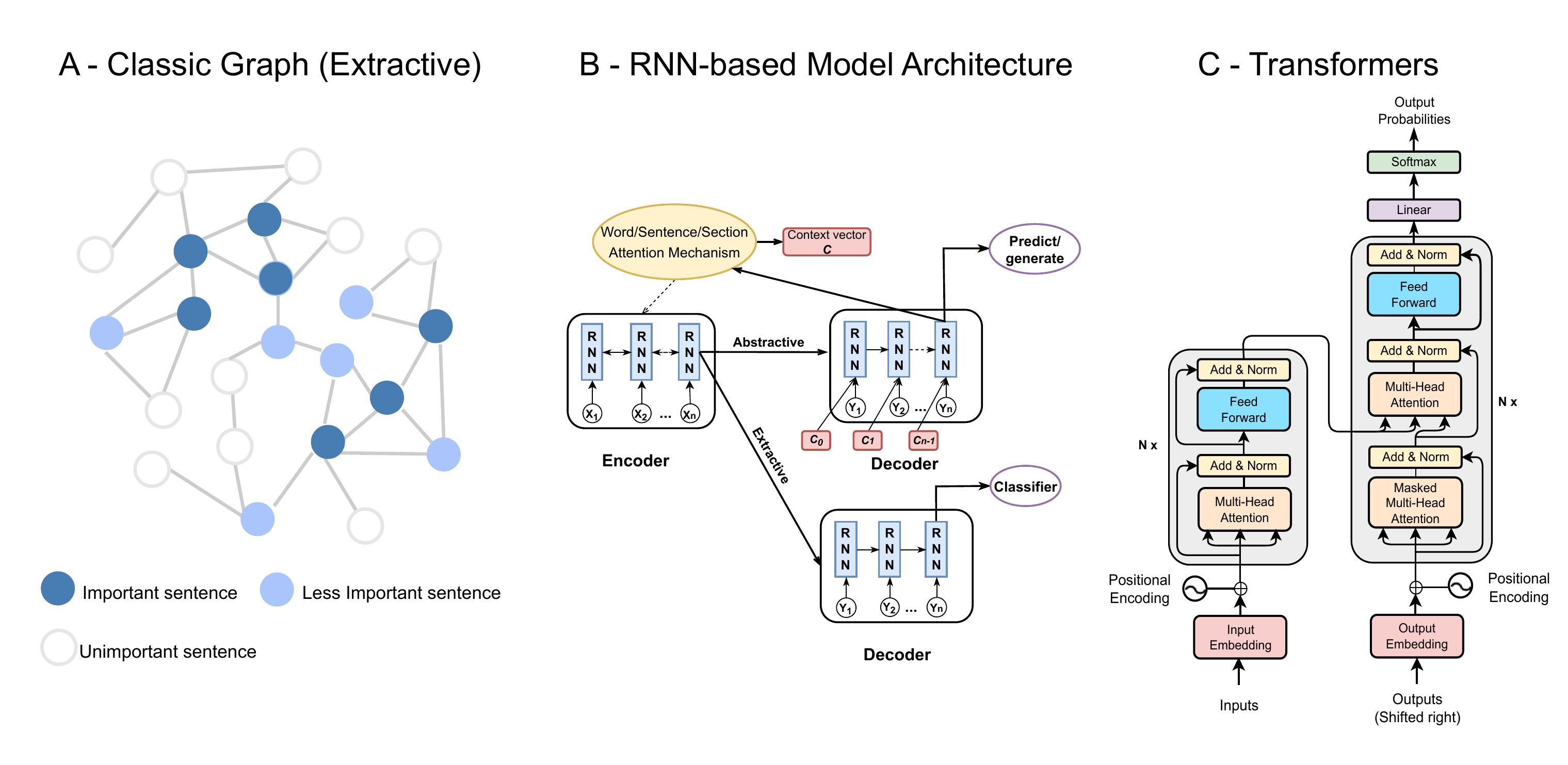}\\
       \caption{Overview of Model Architectures.}
       \label{fig: model_architecture}
\end{figure}

\paragraph{A. Extractive, Abstractive and Hybrid Approach}\,\\ The works in automatic text summarization research are traditionally classified into three different summarization approaches: (i) the extractive approach that involves direct extraction of salient fragments such as sentences of the original documents into a summary \cite{gong2001generic,cheng2016neural}, (ii) the abstractive approach imitates human behavior of paraphrasing important parts of a document into a summary \cite{rush2015neural,see2017get} and (iii) the hybrid approach that attempts to combine the best of both approaches \cite{gidiotis2020divide,manakul2021longspan}. Intuitively, the extractive summarization method is an easier machine learning task and can be thought of as a classification and/or ranking problem of extracting lexical fragment units (e.g., sentences) into a summary. Contrastively, abstractive summarization requires paraphrasing important ideas of a document into a summary either by rearranging words and phrases from original text or contriving novel wordings while maintaining the factual consistency of the generated summary with the original document. 

Since the extractive summarization approach only extracts and arranges the original text that it believes to be salient and does not alter the original text, it enjoys the benefit of generating summaries that are factually consistent with the source document \cite{cui2021sliding}. Nevertheless, as human-based summarization often involves paraphrasing ideas and concepts into shorter, concise sentences, the extracted sentences of this approach often contain redundant and uninformative phrases \cite{grenander2019countering}. While there exist extractive summarization models that break a source document into lower lexical units than sentences (e.g., elementary discourse units) \cite{xu2019discourse}, they are often not applied in the long document summarization domain due to the extreme length of the input document. 

On the other hand, mimicking how humans write summaries, the abstractive summarization approach presents a blue-sky potential of generating summaries that are fluent, concise and relevant to the source document \cite{see2017get}. It can also incorporate external knowledge to the summary depending on the needs of a user \cite{maynez2020faithfulness}. However, at the current stage of development, summaries generated by the state-of-the-art abstractive models often contain a significant amount of content that is factually inconsistent with the source document, limiting its application in commercial settings \cite{cao2018faithful,kryscinski-etal-2020-evaluating}.

Finally, in response to the limitation of current model architectures and designs, the hybrid summarization approach only differs from the abstractive summarization approach in that it takes in a carefully chosen subset of the original input document rather than the entire input document in its original form \cite{pilault2020extractive,gidiotis2020divide}. This extra step reduces the burden on the abstractive summarization models that have to generate an abstract summary and select important content at the same time. This approach is used more often in the long document summarization domain because current models still fail either (a) at reasoning over extremely long texts \cite{meng2021bringing,manakul2021longspan} and/or (b) suffers from memory complexity issues and hardware limitations that prevent it from processing over a long input text \cite{zaheer2020big,huang2021efficient}.

\paragraph{B. General Taxonomy}\,\\
In each long document summarization model, this paper breaks down a model into two different constituents: (i) Main Architecture and (ii) its Mechanisms. The main architecture refers to the core framework structure that a model uses and the mechanisms are the different settings or modifications implemented by a model to the main architecture. Two differing models may use the same main architectures but are implemented with different mechanisms, and vice-versa. For example, models that use graph-based main architecture may use different encoding mechanisms in vectorizing the sentences of an input document. The following describes the various main architectures of the summarization models together with how previous works differ in the mechanisms employed to generate long document summaries.

\subsection{Main Architecture and its Mechanisms}
In the search for optimal architectural settings of summarization systems, the research field started out with many different novel designs of main architectures and mechanisms but often converge towards a few ideas that are often most effective until another ground-breaking idea that leap-frogs the performance of previous systems, and the cycle repeats. 
\,\\\,\\
\textit{1. Graph Architecture:}

For the extractive summarization approach, the classic graph architecture involves a two-stage process of mapping a document into a graph network, where the vertices are sentences and the edges are the similarity between these sentences, and extracting the top-$K$ sentences. The sentences are ranked based on the graph centrality scoring of each sentence \cite{mihalcea2004textrank,erkan2004lexrank}. As there are many different ways to (a) encode or vectorize a sentence before calculating the similarity between them and (b) calculate the centrality score of each sentence, research involving this architecture often differs only in these two mechanisms. For example, with respect to the former mechanism, graph architecture in the past \cite{mihalcea2004textrank,erkan2004lexrank} encodes sentences based on word-occurrence or term frequency-inverse document frequency (Tf-Idf) while graph architecture today \cite{zheng2019sentence,liang2021improving} encodes sentences with state-of-the-art pre-trained models. On the other hand, to improve the centrality scoring mechanism, PacSum \cite{zheng2019sentence} and FAR \cite{liang2021improving} adjust the centrality score of a sentence based on whether the other sentences come before or after it, while HipoRank \cite{dong2021discourse} exploits the discourse structure contained in by adjusting the centrality score with positional and sectional bias. In general form, given a set of sentences in the original source document, $D = \{s_1, s_2,...,s_m\}$ with the inter-sentential similarity relations represented as $e_{ij} = (s_i,s_j) \in E$  where $i \neq j $, the following illustrates the aforementioned architecture in computing the scoring for each sentence: 

$$centrality(s_i) = \sum_{j \in \{1,...,i-1,i+1,...,m\}} e_{ij} * Bias(e_{ij})$$ 

The similarity between each sentence is computed using similarity measures such as dot product or cosine similarity, and the sentences are vectorized using Tf-Idf or BERT representation values. The final summary is generated by extracting the top-k sentences ranked by $centrality(s_i)$. Importantly, while there are other classical architectures \cite{gong2001generic,vanderwende2007beyond}, the graph architecture is worth a separate mentioning here due to the fact that (a) it remains as a strong baseline against other advanced architectures, (b) it can effectively incorporate external knowledge as an inductive bias to the calculation of the importance of a sentence and (c) it achieves state-of-the-art result in long document unsupervised extractive summarization setting when integrated with current state-of-the-art pre-trained models \cite{dong2021discourse,liang2021improving}. Lastly, other than the \hlt{multi-sentence compression approach}
\cite{boudin-morin-2013-keyphrase,ju2020monash,zhao2020summpip} that may be extended to long document summarization tasks, there has been no applicable work on classical graph-based architecture for long document \textit{abstractive} summarization.
\,\\\\
\textit{2. Other Classical Architectures:} 

In the early work of automated, non-neural text summarization models, past research mostly focused on the extractive summarization approach due to the difficulty of the abstractive summarization task. The main architectures that were tested ranged from support vector machines \cite{chali2009svm, shivakumar2015text}, Bayesian classifiers \cite{kupiec1995trainable}, decision trees \cite{mani1998machine,knight2002summarization} to citation network-based summarization \cite{qazvinian2008scientific}. The ones that remained relevant when comparing model performance across various benchmarks in long document summarization settings are LSA \cite{gong2001generic}, which is based on Singular Value Decomposition (SVD), and SumBasic \cite{vanderwende2007beyond} that ranks sentences by simple average word-occurrence probability \cite{vanderwende2007beyond}. 
\,\\\\
\textit{3. Recurrent Neural Networks:}

Extractive summarization that employed neural networks with continuous representations rather than pre-trained word embeddings on traditional techniques \cite{kobayashi2015summarization,yogatama2015extractive} was proposed by \citet{cheng2016neural}. The model implemented an RNN encoder-decoder architecture with attention mechanism to locate the region of focus during sentence extraction process. Nevertheless, due to the lack of a large-scale long document dataset and the RNN's inability in capturing long-range temporal dependencies across a long input text, it wasn't until \citet{xiao2019extractive} that tried implementing  LSTM-minus (a variant of RNN) on solving long document summarization task. Typical of a long document summarization system, it incorporates discourse-information (i.e. section structure) of the source document by encoding the section-level and document-level representation into each sentence to significantly boost the model performance. \citet{pilault2020extractive} also suggested two different variants of RNNs on extractive summarization for long document summarization. Rather than utilizing pre-trained word embeddings, they implemented a hierarchical LSTM to encode words and sentences separately.

When it comes to the abstractive summarization approach, \citet{celikyilmaz2018deep} proposed multiple communicating agents to address the task of long document summarization. However, as compared to other simpler architecture, this approach did not gain significant traction after the introduction of the first large-scale dataset on long scientific documents. Together with the contribution of two most commonly used long scientific document datasets, arXiv and PubMed, \citet{cohan2018discourse} presented an LSTM encoder-decoder architecture where the decoder attends to each section of the source document to determine section-level attention weights before attending to each word. While similar architectures have been widely used in prior works, this work effectively incorporates discourse information that suits the long document summarization task well.
\,\\\\
\textit{4. Transformers: }

The Transformer model proposed in 2017 together with the pre-trained Bidirectional Encoder Representation from Transformers (BERT) model that was based on the Transformer model itself have taken the NLP area by storm \cite{vaswani2017attention,devlin2019bert}. Like other NLP tasks, subsequent summarization model architectures have changed significantly to take advantage of these two momentous ideas. Importantly, BERTSum \cite{liu2019text} showed that by modifying the BERT-segmentation embeddings, it can capture not only sentence-pair inputs but multi-sentential inputs. The BERTSum model could effectively solve both extractive and abstractive summarization tasks. The extractive summarization model proposed by BERTSum involves stacking a Transformer-based classifier on top of the fine-tuned BERT to select and extract salient sentences while the abstractive summarization model involves a classic encoder-decoder Transformer framework where the encoder is a fine-tuned BERT and the decoder is a randomly-initialized Transformer that is jointly trained together in an end-to-end manner. While effective, this architecture cannot be implemented to solve long document summarization tasks due to the BERT's input length limit of 512 tokens. In this survey, we define Transformer as the main architecture and settings related to the Transformer model as the mechanisms including the use of different types of pre-trained models. As Transformer has effectively replaced most main architectures in both summarization approaches as state-of-the-art models, the various mechanisms applied in the long document summarization context will be thoroughly discussed in section 4.3.

\subsection{Mechanisms of Transformer-based Architectures}\label{subsec: transformer_mechanism}

\begin{figure} [h!]
    \centering
       \includegraphics[width=0.77\textwidth]{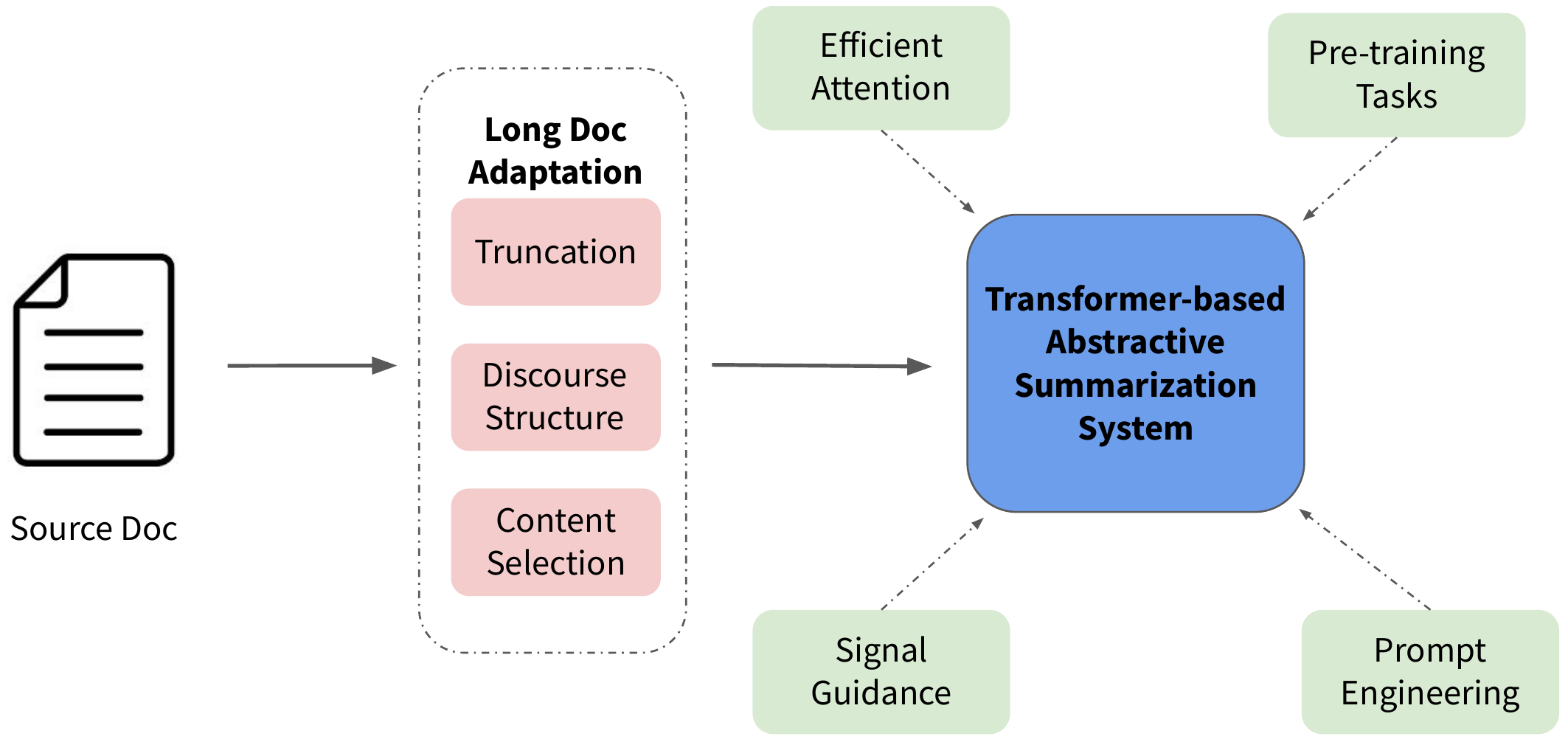}\\
       \caption{Overview of Transformer-based Abstractive \& Hybrid Summarization Models.}
       \label{fig. transformer_mechanisms}
\end{figure}

Transformer-based model is ubiquitously state-of-the-art across a wide range of tasks in the NLP domain. In line with this development, recent works in the long document summarization models often involve using the same Transformer base architecture but with different proposing mechanisms. These Transformer-based models involve implementing novel mechanisms with long document adaptations to ensure the task of summarizing a document with significantly longer input sequence texts can be effectively addressed. The mechanisms used by extractive, abstractive and hybrid Transformer-based summarization models are described in the following with an overview of mechanisms used by abstractive and hybrid summarization models shown in Figure \ref{fig. transformer_mechanisms}. 

\paragraph{A. Extractive Transformer}\,\\
As Transformer and its pre-trained models are optimized for short document settings, they may not reason well over long text sequences if not properly fine-tuned. To this end, \citet{cui2020enhancing} proposed combining neural topic modeling together with BERT in learning a topic-enhanced, inter-sentence relationship across the entire document. Nonetheless, the issues of memory complexity and input token length limits were not resolved and significant source text is truncated under this research setting. Recently, \citet{cui2021sliding} proposed a memory network that incorporates graph attention networks and gated recurrent units to dynamically select important sentences through sliding a window along the entire source document. This approach can effectively integrate the pre-trained BERT model for long document summarization task by limiting its usage within each window, where the window size is set to be lower than or equal to 512 tokens. 
\paragraph{B. Abstractive Transformer}

\,\\\textit{1) General Sequence-to-Sequence Pre-training Task}\\
Since the advent of BERT \cite{devlin2019bert}, various large-scale models with different pre-training tasks have been introduced. As summarization with the abstractive approach is naturally a sequence-to-sequence task, a pre-trained model with a sequence-to-sequence objective task would suit it better rather than an encoder-only (e.g., BERT/RoBERTa) or a decoder-only (e.g., GPT-2/GPT-3) pre-trained model. In the summarization domain, Bidirectional and Auto-Regressive Transformers (BART) \cite{lewis2020bart} and Text-to-Text Transfer Transformer (T5) \cite{raffel2020exploring} are the two most widely used sequence-to-sequence pre-trained models. BART is pre-trained on a self-supervised task of reconstructing arbitrarily corrupted text while T5 is pre-trained on both unsupervised and supervised objectives, such as token masking, as well as translation and summarization. Interestingly, none of the supervised Transformer models in the long document summarization domain has implemented a summarizer with T5 pre-training task despite its success in the short document domain \cite{rothe-etal-2021-thorough}. 
\\
\\
\textit{2) Gap-Sentence Generation (GSG) Pre-training Task}\\
Other than the generalized pre-training task like BART and T5, PEGASUS \cite{zhang2020pegasus} attempted to significantly advance the progress in the abstractive summarization field through large-scale pre-training with objectives that are specific to the summarization task. The proposed model is self-trained on two large scale datasets (C4 and HugeNews) with the gap-sentence generation pretraining task. Gap-sentence generation pretraining task draws a close resemblance with the general summarization task by self-supervising the model to generate sentences that are masked entirely in a given document. At the time of PEGASUS model release, the model effectively achieves state-of-the-art results across 12 different benchmark datasets, including long document arXiv and PubMed dataset. 
\\
\\
\textit{3) Efficient Attentions} \\
The vanilla Transformer models that utilize full attention have a memory complexity \(O(n^2)\). This attribute limits its wider usage across many domains, including long document summarization. For example, to circumvent the input tokens limits of PEGASUS, DANCER \cite{gidiotis2020divide} summarizes each section of the long document separately and concatenates each of them to form the final summary. As not all benchmark datasets contain discourse information such as section structures, this limits the model usage in many long document summarization settings. To this end, researchers have proposed various ingenious ideas to reduce the memory and time complexity of Transformer models. The variants of Transformer models that require less memory are often known as efficient Transformers \cite{tay2020efficient,tay2020long} and the mechanism is referred to as efficient attentions \cite{huang2021efficient}.

Longformer \cite{beltagy2020longformer} combines local attention, stride patterns and global memory for fine-tuning pre-trained BART to effectively summarize long documents with a maximum input length of 16,384 tokens as opposed to the 1,024 token limit of the original BART model. The model achieved state-of-the-art results in the long document summarization along with other NLP tasks when the model was introduced. BigBird \cite{zaheer2020big} also implemented the efficient attention mechanism on Transformer-based abstractive summarizer by utilizing the same attention modifications as Longformer with an additional random pattern to achieve matching performance results in terms of ROUGE score. An important work by \citet{huang2021efficient} explores and compares the performance of different variants of efficient transformers in the context of long document summarization. 

\begin{figure*} []
    \centering
       \includegraphics[width=0.8\textwidth]{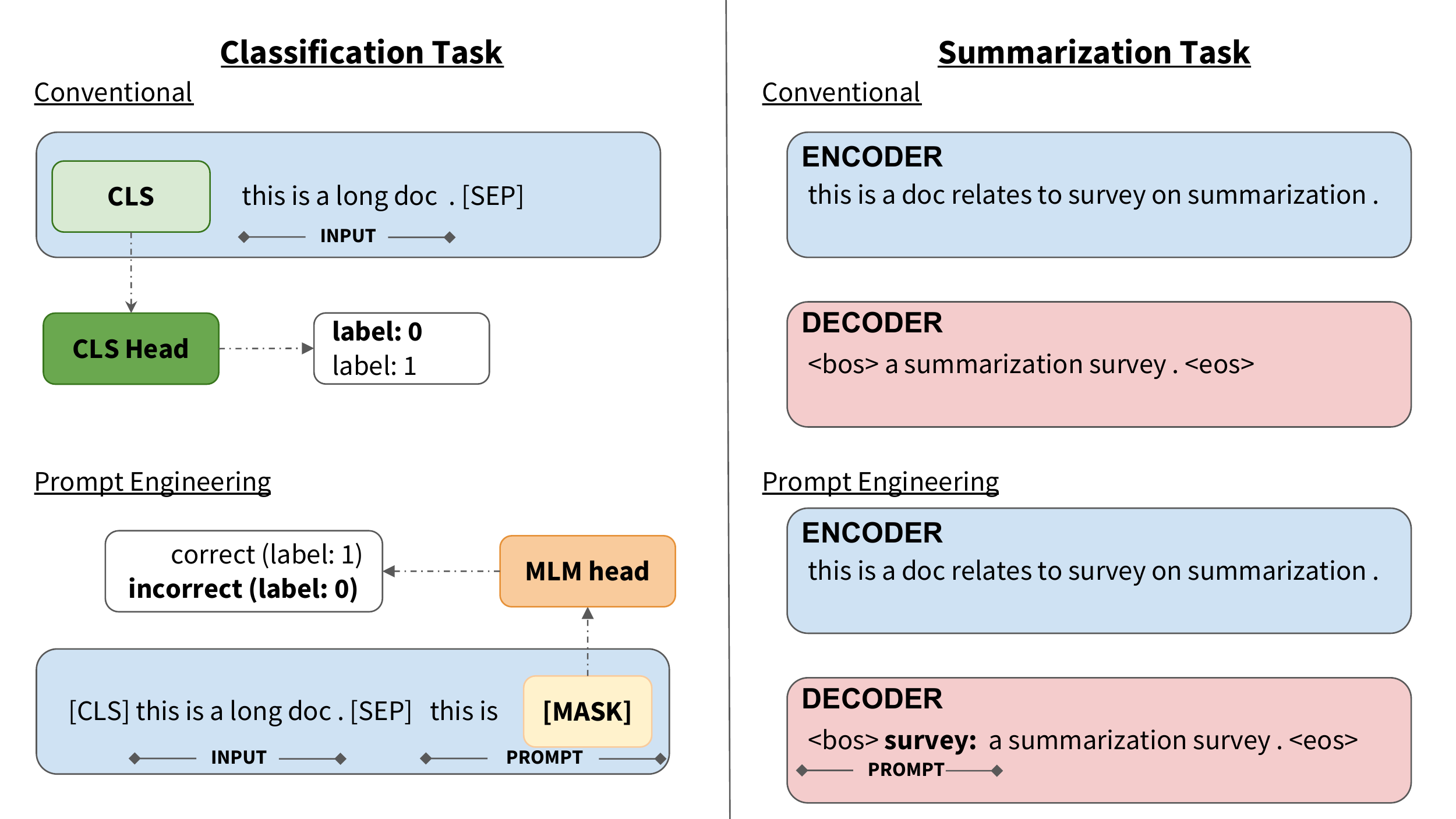}\\
       \caption{How prompt engineered language tasks differ. Classification task example from \cite{gao-etal-2021-making} and summarization task using keywords from original document as language prompt from \cite{he2020ctrlsum}.}
       \label{fig. prompt_mechanisms}
\end{figure*}
\,
\\
\textit{4) Prompt-Engineering}\\
The GPT-3 model \cite{brown2020language} has strongly demonstrated that large-scale pre-trained language models can achieve impressive results on numerous downstream language tasks in zero- and few-shots experimental settings. Rather than fine-tuning the language models for specific tasks in a conventional way, natural language prompts and task demonstrations were created for GPT-3 to infer and complete the tasks. Importantly, this is different from the conventional tagging method such as <bos> token for conditional generations or taking [CLS] tag from BERT for classification tasks. Prompt engineering refers to taking the extra step to design a natural language prompt or template that can optimize the pre-trained model for a specific task. Figure \ref{fig. prompt_mechanisms} illustrates this important difference. 

Many works have explored the ways of uncovering the right prompt for various downstream language tasks to significantly boost the performance of pre-trained models \cite{gao-etal-2021-making,talmor2020olmpics}. In the long document summarization research area, CRTLSum \cite{he2020ctrlsum} achieves significant improvement on vanilla fine-tuned BART model on arXiv dataset through prompt-engineering. The model attempts to more effectively use the pre-trained BART model with the help of extracted keyword prompts, as shown in Figure \ref{fig. prompt_mechanisms}. Further, the work also showed that, given an optimized language prompt, the implemented BART summarization model can achieve ROUGE score that matches ROUGE score of oracle summaries in the test dataset.
\\\\
\textit{5) Signal Guidance}\\
Unlike a prompt-engineering mechanism that requires an engineered language prompt or template for a given task, the signal guidance mechanism relates to utilizing signals as inputs to lead models in better identifying and summarizing important content of source texts. Using this approach, the GSum \cite{dou2021gsum} model implemented a fine-tuned BART model with dual encoders, one for input document and another for extracted signals, and a decoder that attends to both encoded representations. Similar signal based approach is also used by \citet{ju2021leveraging} to implement an unsupervised pipeline-based long document summarization model.
\\
\\
\textit{6) Discourse Bias}\\
Similar to the signal guidance mechanism, discourse bias involves the inclusion of the discourse structure of a source document such as the section of a sentence as signals for summarization systems to better identify and summarize important content in the original source text \cite{christensen2013towards,yasunaga2017graph}. This mechanism can be classified under signal guidance but is mentioned separately due to its effectiveness and popularity in Transformer \cite{pilault2020extractive,gidiotis2020divide} and non-Transformer \cite{cohan2018discourse,dong2021discourse} based long document summarization models. Unlike short documents, long documents often contain discourse structure information such as table of content, section structures, references and others to guide a human reader in comprehending the original document and previous works have exploited this information to achieve state-of-the-art results. Nonetheless, previous works in the long document summarization domain only utilized discourse information that is made available by the benchmark datasets and did not implement automatic discourse parsing using RST trees \cite{mann1988rhetorical} and coreference mentions due to the difficulties of building an effective representation for a document with extreme input length \cite{ji2014representation,xu2019discourse}. 

\paragraph{C. Hybrid Transformer - Content Selection + Abstractive Summarization}\,\\
The hybrid summarization approach only differs from the abstractive approach in that it takes in a carefully chosen subset of the input document rather than the entire input document. This extra step reduces the burden on the neural summarizers that have to generate an abstract summary and select important content at the same time. Some also refer models that utilize the hybrid approach as retrieve-the-summarize model because it involves retrieving a subset of long document text before summarizing it \cite{zhang2021exploratory}. TLM+Ext \cite{pilault2020extractive} first implemented this method by limiting inputs of the scientific articles in arXiv datasets as the introduction of the document, a subset of carefully selected sentences of the original article using extractive summarization approach, and, finally, include the remaining text if there remains extra space for Transformer-based decoder. However, given the effectiveness of sequence-to-sequence neural models, one limitation of this work is that it only utilizes a decoder framework to generate the final summary rather than an encoder-decoder framework that most subsequent works on abstractive and hybrid summarization approach do. Consequently, LoBART \cite{manakul2021longspan}
proposes a hybrid summarization system that completes a summary generation in two separate steps, (i) content selection: using a multi-task RNN, select salient content from the original source document until the total text output reaches the limit of the sequence-to-sequence pre-trained BART model and (ii) abstractive summarization: summarize the carefully selected subset using a pre-trained BART model with efficient transformer mechanism. SEAL \cite{zhao2020seal} presents a generalized encoder-decoder framework for transformer-based long document summarization and proposed an abstractive summarization system that selects salient content and dynamically choose segments of the selected content for the decoder to attend and summarize in an end-to-end manner. The architecture, however, did not attempt in exploiting the large-scale pre-trained models that were used in most summarization research works. Lastly, facing a similar issue, development in the open-domain question-answering and knowledge-intensive language tasks reflect an interesting parallel with the progress in the long document summarization domain \cite{lewis2020retrieval,petroni2021kilt}.

\subsection{Summary of Trends in Long Document Summarization Systems}
\begin{table*}[!ht]
\resizebox{\textwidth}{!}{%
\small
\begin{tabular}{c|c|c|c|c|c}
\thickhline
\multicolumn{1}{l|}{} &
  \textbf{Model} &
  \multicolumn{1}{c|}{\textbf{Architecture}} &
  \textbf{Pre-Train} &
  \textbf{Long Document Mechanism} &
  \textbf{Max Token} \\ [0.3ex]
  \hline
\multirow{4}{*}{\textit{\begin{tabular}[c]{@{}c@{}}Unsupervised \\ Extractive\end{tabular}}}
& PacSum \cite{zheng2019sentence} & Graph & BERT & Discourse Bias & - \\
 & HipoRank \cite{dong2021discourse}      & Graph                 & BERT & Discourse Bias & -    \\
 & FAR \cite{liang2021improving} & Graph & BERT & Facet-Aware Scoring & - \\
 & IBSumm \cite{ju2021leveraging}  & Pipeline &  SciBERT  & Signal Guidance & - \\ [0.2ex]
 \hline
\multirow{4}{*}{\textit{\begin{tabular}[c]{@{}c@{}}Supervised \\ Extractive\end{tabular}}} 
 & GlobalLocal \cite{xiao2019extractive}   & RNN  & - & Discourse Bias    \\
 & Sent-CLF/PTR \cite{pilault2020extractive} &
  RNN & - &
  - &
  - \\
 & Topic-GraphSum \cite{cui2020enhancing} & GAT                    & BERT & Neural Topic Modelling       & -  \\
 & SSM-DM  \cite{cui2021sliding} & DMN         & BERT & -       & -    \\ [0.2ex] 
\hline
\multirow{7}{*}{\textit{\begin{tabular}[c]{@{}c@{}}Supervised \\ Abstractive\end{tabular}}} &
  Discourse-Aware \cite{cohan2018discourse} &
  RNN & - & Discourse Bias &
  - \\
 & Longformer  \cite{beltagy2020longformer}   & Transformer                          &  BART  & Efficient Attention       & 16,384     \\
  & BigBird \cite{zaheer2020big}       & Transformer  & PEGASUS              & Efficient Attention       & 4,096 \\
    & GSUM \cite{dou2021gsum}       & Transformer  & BART             & Signal Guidance       & 4,096 \\
 & CRTLSum \cite{he2020ctrlsum} & Transformer & BART                                  & Prompt Engineering       & 1,024 \\
  & HAT-BART \cite{rohde2021hierarchical} & Transformer & BART & Hierarchical Attention & 1,024 \\
 & HEPOS \cite{huang2021efficient} & Transformer   & BART                & Efficient Attention     & 10,240  \\ \hline
 \multirow{4}{*}{\textit{\begin{tabular}[c]{@{}c@{}}Supervised \\ Hybrid\end{tabular}}} &
TLM+Ext \cite{pilault2020extractive}  & Transformer &  -       & Content Selection + Discourse Bias       & -    \\ 
 & DANCER   \cite{gidiotis2020divide} & Transformer   &  PEGASUS                           & Content Selection + Discourse Bias & -  \\
 & SEAL \cite{zhao2020seal}  & Transformer & -  & Content Selection w/ Segment-wise Scorer       & -    \\
& LoBART \cite{manakul2021longspan} & Transformer & BART & Content Selection + Efficient Attention & - \\ [0.2ex]
\thickhline
\end{tabular}%
}
\caption{ Long Document Summarization Models in Chronological Order. Max token represents the maximum input sequence length that the model can process and any text that exceeds this cutoff point will be truncated.}
\label{tab. lds_history}
\end{table*}

Table \ref{tab. lds_history} summarizes the trends and developments in long document summarization models as discussed above. The two standout base architectures that are used in the long document summarization domains are graph-based ranking algorithm for unsupervised extractive models and pre-trained Transformer for supervised abstractive models. While both architectures were initially proposed and tested on short documents, they can be effectively adapted to summarize long documents after incorporating novel mechanisms.\footnote{For a brief description of each model, please refer to the Supplementary Materials.}

\textbf{Finding 1. Graph-based Extractive Models with Discourse Bias:} Classical graph-based unsupervised extractive models have been found to suffer from picking similar sentences that results in a summary with redundant sentences \cite{liang2021improving}. To this end, HipoRank \cite{dong2021discourse} implements the graph-based architecture for unsupervised extractive summarization by including the sectional information of ArXiv/PubMed as inductive bias when calculating centrality scoring to achieve state-of-the-art results. The discourse bias mechanism is commonly incorporated by other proposed summarization models, including models with RNN and Transformer base architectures. 

\textbf{Finding 2. Pre-training task for Abstractive Summarization Models:} Interestingly, despite having other pre-trained sequence-to-sequence models such as T5 \cite{raffel2020exploring}, BART and PEGASUS are the only two Transformer-based pre-trained models that were used for long document summarization \cite{lewis2020bart,zhang2020pegasus}. Nevertheless, as both pre-trained models are trained on short documents, they have an input limit of 1,024 tokens. To process long documents that are longer than this limit, the pre-trained Transformers will have to incorporate long document mechanisms to extend the input limits.

\textbf{Finding 3. Long Document Mechanisms for Transformer:} As pre-trained models were often trained on large-scale datasets with input limit length between 512 to 1,024 \cite{devlin2019bert,lewis2020bart}, these Transformer-based pre-trained models were optimized for short document language tasks rather than long documents. Without any long document mechanisms to adapt these models for the long document summarization task, \citet{meng2021bringing} has shown that BART cannot summarize a long document effectively. Other than the discourse bias mechanism, we observe that (a) efficient attention and (b) content selection mechanisms are the two most notable long document mechanisms. As the content selection mechanism requires a separate retriever to extract salient content from the source (i.e., the hybrid approach), we distinguish Transformer models with content selection mechanism as the retrieve-then-summarize model \cite{zhang2021exploratory} and the pure encoder-decoder Transformer without this mechanism as an end-to-end model for the rest of this work. Lastly, it is also important to note that both mechanisms can be jointly implemented within a single architecture, where the content selection mechanism will extract a longer subset of input to be processed by a Transformer with efficient attention \cite{manakul2021longspan}.
\section{Multi-dimensional Analysis of Long Document Summarizers}
Given the important findings in the graph-based unsupervised extractive model and the Transformer-based supervised abstractive model in the previous section, we design an experiment with the aim of thoroughly understanding the reasons behind the popularity of these architectures and its mechanisms. Our experiment tests out the graph-based extractive and Transformer-based abstractive summarization architectures and its mechanisms on the arXiv benchmark dataset. The documents in the arXiv dataset have an average length of 6,446 tokens. Mechanisms of the supervised extractive approach are not examined as the architectures used between the proposed models vary greatly.

\subsection{Implementation}

\subsubsection{Graph - Unsupervised Extractive}\,\\
To investigate the effect of incorporating long document discourse structure information, we experiment with four unsupervised graph models by varying the two following mechanisms:
\begin{description}
  \item[$\bullet$ Sentence Encoder Mechanism:] Sentences of source text are encoded either using Term frequency–inverse document frequency (Tf-Idf) \cite{jones1972statistical} or BERT SentenceTransformer \cite{reimers2019sentence}. 
  \item[$\bullet$ \hlt{Discourse} Bias Mechanism:] For both models that implement Tf-Idf or BERT sentence encoder, we experiment a model with the long document \hlt{discourse} bias mechanism and one without the bias. \hlt{For models without a discourse bias mechanism, the centrality score of each sentence is computed based on the summation of cosine similarity between other sentences. For models with a discourse bias mechanism, the mechanism implemented follows the work of \citet{dong2021discourse}. For each sentence, we adjust centrality score based on the sentence position within each section, $c_{intra}(s_i)$, and the sentence's section position within the document, $c_{inter}(s_i)$. Sentences that are closer to the section and document boundaries will be given higher importance. The "discourse-aware" centrality score for each sentence is: 
$$c(s_i) = \mu_1 \cdot c_{inter}(s_i) + c_{intra}(s_i)$$
where $\mu_1$ is a weighting factor for inter-section centrality. Following the original author, we fine-tune the weighting factor based on validation set. To ensure comparability, the maximum length of summary for all unsupervised extractive models is set to be 242 tokens\footnote{Details of implementation are reported in the Supplementary Materials.}.} 
\end{description}


\subsubsection{Transformer - Supervised Abstractive}\,\\
To study the current state-of-the-art abstractive neural summarizers, we experiment with two different pre-trained Transformers, BART and PEGASUS. For BART, we analyze the long document mechanisms from the perspective of two common approaches:

\begin{description}
  \item[$\bullet$ End-to-End:] We experiment with three end-to-end BART models. Firstly, a vanilla BART model with full self-attention that will truncate any input text that exceed 1,024 tokens. Then, two BART models with efficient longformer attention \cite{beltagy2020longformer} that can extend up to 4,096 and 16,384 input tokens respectively. The main goal of assessing end-to-end BART with and without efficient attention is to assess how the quality of the generated summary is affected when BART is adapted from a short document summarizer into long document summarizer by allowing it to process long input sequences at the cost of full self-attention.  For implementation, due to lack of computational resources, we only fine-tuned original BART on arXiv and obtain the weights of longformer that was trained on arXiv from the original author\footnote{https://huggingface.co/allenai/led-large-16384-arxiv}. 
  \item[$\bullet$ Retrieve-then-summarize] For BART retrieve-then-summarize model, our experiment follows entirely the implementation of LoBART by \cite{manakul2021longspan}\footnote{https://github.com/potsawee/longsum0}. LoBART has two variants: (a) BART with full self-attention that takes in a selected subset of input text with a maximum length of 1,024 tokens and (b) BART with efficient local attention where maximum length of subset input is 4,096 tokens. The only difference between the two variants is the amount of content to be retrieved from the original source text before feeding it into the Transformer BART model. The two main objectives of experimenting with the retrieve-then-summarize BART models are to assess: (1) the effectiveness of content selection mechanism in adapting short document BART to summarize a long document and (2) whether the performance is improved when content selection mechanism is combined with efficient attention mechanism. 
\end{description}

As there is no existing framework that applies content selection mechanism on PEGASUS, we only experiment two end-to-end PEGASUS model: original PEGASUS\footnote{https://github.com/google-research/pegasus} and PEGASUS with BigBird efficient attention\footnote{https://github.com/google-research/bigbird}. The models have input token limit of 1,024 and 4,096 respectively. Pre-trained weights for both model variants on arXiv benchmark are obtained directly from the original author. 

\subsubsection{Assessing Model Outputs}\,\\
Rather than relying entirely on ROUGE score like most summarization research settings, we use four different metrics to analyse model summary outputs from three important dimensions (D\#): relevance, informativeness and semantic coherence.

\begin{description}
  \item[$\bullet$ D1 - Relevance] of a summary is the extent to which a summary contains the main ideas of a source. We use ROUGE \cite{lin2004rouge} and BERTScore \cite{zhang2019bertscore} metrics to measure relevancy of the candidate summary.
  \item[$\bullet$ D2 - Informativeness] is the amount of new information and knowledge a summary brings to the reader \cite{peyrard2019simple}. This information may not necessarily be key to the narrative of the source but should add value to readers. For example, limitations of an academic article are not central to the narrative but do add value to readers. This metric tests a model architecture's ability to effectively generate summary that can cover different aspects of the original source text. This is approximated by the percentage of sections that are covered by a candidate summary, where we assume each sentence of the candidate summary covers a particular section and the sentence belongs to the section where it achieves the highest ROUGE-L score. 
  \item[$\bullet$ D3 - Semantic Coherence] measures whether a summary is fluent and semantically coherent. Following \citet{bommasani2020intrinsic}'s implementation, this is approximated as: 
$$SC(S)= \frac{\sum_{i=2}^{||S||} NSP(s_i|s_{i-1})}{||S||} $$ where $NSP(.)$ is BERT NSP function, and $s_i$ denotes the position of a sentence in the candidate summary. However, \citet{bommasani2020intrinsic} did not fine-tune the general pre-trained BERT model while our BERT NSP model is fine-tuned on arXiv using positive and negative sentence-pairs with a final F1-score of 0.92.
\end{description}

\subsubsection{Other Implementation Details}\,\\
For all the model variants implemented, the train, validation and test sample split on the arXiv benchmark dataset are 203,037/6,436/6,440, which is the same for all prior works as they follow the same configuration by the original author \cite{cohan2018discourse}. To ensure consistent preprocessing pipeline, we follow pre-processing of LoBART in all model implementations \cite{manakul2021longspan}. All models that require fine-tuning are trained on the same RTX 3090 GPU with 24 GiB of GPU memory. We use pyrouge package for ROUGE metric. 

\subsection{Results and Analysis}

\begin{table*}[!ht]
\resizebox{\textwidth}{!}{%
\small
\begin{tabular}{lcccccc}
\thickhline 
\thickhline
 \multirow{2}{*}{\textbf{Architecture and its Mechanisms}}&\multicolumn{4}{c}{\textbf{Relevance}}&\multirow{2}{*}{\textbf{Informativeness}}&\multirow{2}{*}{\textbf{Semantic Coherence}} \\
 &\textbf{R-1}&\textbf{R-2}&\textbf{R-L}&\textbf{BERTScore}&&\\ [0.3ex]
\thickhline
\multicolumn{7}{c}{\textit{\textbf{Unsupervised Extractive} - Graph}}\\ [0.5ex]
Tf-Idf & 0.344 & 0.095 & 0.285 & \textbf{0.822} & 0.355  & \underline{0.779} \\
BERT & 0.351 & 0.098  & 0.290  & 0.819  & 0.398 & \textbf{0.792} \\
Tf-Idf + \hlt{Discourse Bias}  & \underline{0.357}   & \textbf{0.113}  & \underline{0.311} & 0.820  & \underline{0.447} & 0.677 \\
BERT + \hlt{Discourse Bias} & \textbf{0.361} & \underline{0.112} & \textbf{0.322} & \underline{0.822} & \textbf{0.513} & 0.667 \\ [0.5ex]
\hline 
\multicolumn{7}{c}{\textit{\textbf{Supervised Abstractive} - BART Variants}}\\ [0.5ex]
\textit{\textbf{End-to-End (Max Token)}}& && & &&\\ 
BART-only (1,024)                                                          & 0.413      & 0.153                & 0.368                & 0.846                 & 0.331                          & 0.851                             \\
BART+LongformerAttn (4,096)                                            & 0.463       & 0.187                & 0.412                & 0.852                 & 0.361                          & 0.855                             \\
BART+LongformerAttn (16,384) &  0.467      & \underline{0.196}                 &    0.418            &  \textbf{0.865}                &              \textbf{0.433}             &     \textbf{0.877}                        \\ [0.5ex]

\textit{\textbf{Retrieve-then-summarize}}& && & &&\\ 
BART+CS                                                             & \underline{0.472}       & 0.193                & \underline{0.419}                & 0.837                 & 0.403                          & 0.812                             \\
BART+LocalAttn+CS                                         & \textbf{0.486}       & \textbf{0.201}                & \textbf{0.422}                & 0.845                 & \underline{0.427}                          & 0.835                             \\ [0.5ex]
\multicolumn{7}{c}{\textit{\textbf{Supervised Abstractive} - PEGASUS Variants}}\\ 
\textit{\textbf{End-to-End (Max Token)}}& && & &&\\ 
PEGASUS (1,024)                                                        & 0.439       & 0.171                & 0.381                & \underline{0.857}                 & 0.333                          & 0.863                             \\
PEGASUS+BigBirdAttn (4,096)                                         & 0.462       & 0.190                & 0.415                & 0.855                 & 0.359                          & 
\underline{0.869}                             \\
\thickhline 
\thickhline 
\end{tabular}%
}
\caption{Experimental Results of Graph-based Unsupervised Extractive and Transformer-based Supervised Abstractive. The best results are in boldface, and the second highest scores are \underline{underlined}. Max token represents the maximum input length where texts that exceed this cutoff point are truncated. \hlt{LocalAttn represents local attention where each token only attends to its neighbouring 1,024 tokens. LongformerAttn and BigBirdAttn represents efficient attention variants proposed by Longformer \cite{beltagy2020longformer} and BigBird \cite{zaheer2020big}.}}
\label{tab. result}
\end{table*}

The following discusses the experimental \textbf{findings} based on the results shown in Table \ref{tab. result}.

\subsubsection{Graph - Unsupervised Extractive}

\,\\

\textbf{Finding 1. Global Word Representation versus Contextual Embedding:} Using BERT as the sentence encoder mechanism boosts the unsupervised summarization model performance in the relevancy and informativeness dimension. We hypothesize this is due to the semantic reasoning capability of BERT in encoding important sentences that are informative but are not worded in the same way as the other important sentences when similarity and centrality scoring are computed. This is particularly important for long document as it has higher compression ratio and a higher chance of having sentences with the exact same information being repeated multiple times in the source text. Thus, not encoding the sentences with semantically rich encoding may result in a summary with more redundancy.

\textbf{Finding 2. Discourse Bias Mechanism boosts Relevance and Informativeness:} The inclusion of positional and sectional bias when computing sentence centrality score greatly improves the architecture's ability in capturing more relevant sentences in the long document and generating a more informative summary, validating our hypothesis in section 4. However, since extractive models merely combine the extracted sentences, these sentences that come from various sections likely caused a drop in the semantic coherence of the final summary outputs. 

\subsubsection{Transformer - Supervised Abstractive}

\,\\

\textbf{Finding 1. Diminishing Return of \hlt{PEGASUS} Pre-training:} As compared to the BART-only model, the PEGASUS-only model achieved greater performance across all dimensions, indicating that \hlt{PEGASUS pre-training mechanism} helps a Transformer-based model in writing more relevant, informative and semantically coherent summaries. \hlt{As PEGASUS is pre-trained on a different corpus as compared to BART \cite{lewis2020bart,zhang2020pegasus}, it is not conclusive whether the GSG pre-training task and/or the difference in the pre-training corpus contributed to the superior performance of PEGASUS. Interestingly, the performance gain} is not as obvious when the input sequence length is allowed to be extended to 4,096 with efficient attention. This could be due to the difference in the efficient attention mechanism used or the need for predicting salient content outside the truncated text has diminished.

\textbf{Finding 2. Mixed Results on Retrieve-then-summarize Models:} For both retrieve-then-summarize BART models, we see state-of-the-art result achieved in terms of the standard ROUGE score metric. The result showed improvement across all dimensions when the Transformer model processed a longer subset extracted by the retriever, demonstrating the effectiveness of Transformer in computing pairwise relations between tokens to identify salient content. Except for the BART with longformer attention (16,384), retrieve-then-summarize models also performed better in informativeness as the models are allowed to process the entire source documents in arXiv dataset. However, the retrieve-then-summarize models performed the worst in semantic coherence when compared to the other abstractive summarization models. As the content selection mechanism is not trained in an end-to-end manner, we hypothesize that this is due to the inevitable disconnect between the content selection mechanism and the encoder-decoder Transformer model at the inference stage. Further, it is possible when the retrieved subset extracted by the content selection mechanism are not ordered in its original form, the incoherence of the subset cascade downwards to the final summary output, causing a drop in semantic coherence. This finding also illustrates the importance of measuring model performance in a multi-dimensional way rather than relying entirely on ROUGE score that has found to have important limitations \cite{kryscinski2019neural,bhandari2020re}. 

\textbf{Finding 3. Transformer's Reasoning Capability over Long Sequences:}
Holding the pre-trained BART model constant, extending the total input token limits for the pre-trained Transformer improves the summarizer's ability in generating a summary that is more relevant, informative and semantically coherent. This finding is consistent with a human evaluation experiment by \cite{huang2021efficient}, providing confidence to the automatic evaluation metrics used in this work. The impact of processing only 1,024 tokens is particularly obvious when it comes to the informativeness of the summary output where BART (1,024) informativeness score is 10 points lower than BART (16,384). Importantly, ROUGE score again did not fully capture this performance difference, highlighting the limitation of traditional summarization research setting of measuring model performance using only ROUGE score. Lastly, this finding suggests that unlike the result of \citet{meng2021bringing}, our experiment demonstrates that Transformer can reason over long sequences given that the right configuration is made to fine-tuned the model for specific downstream task. 
\\\\
Through ad-hoc experiment, we systematically analyze the common approaches in long document summarization domain. The experimental result demonstrated that exploiting explicit discourse structures of long documents in unsupervised models and processing longer inputs with long document adaptation on pre-trained Transformer models can yield promising outcomes for the long document summarization task. The result also showed that retrieve-then-summarize model can achieve state-of-the-art results in terms of ROUGE score but may generate less coherent summaries.

\subsection{Limitation of Experiment}
Recent studies have found that summary outputs of state-of-the-art abstractive summarization models contain factual inconsistency in up to 30\% of summary output \cite{cao2018faithful,kryscinski-etal-2020-evaluating,maynez2020faithfulness}. To address the aforementioned issues, various models and metrics have been proposed to measure the factual consistency of candidate summaries conditioned on the source documents \cite{goyal-durrett-2020-evaluating,durmus2020feqa}. Nonetheless, due to the limitations of the proposed metrics including the input length limit of pre-trained models, difficulty of implementation and performance variation across benchmarks \cite{pagnoni-etal-2021-understanding}, we did not measure the factual consistency of summary outputs and represents an important limitation of the multi-dimensional analysis experiment above. This is despite after trying out various adaptations on the textual entailment approach proposed by \citet{maynez2020faithfulness}, our tested models have almost no discriminative ability and was thus not used. The robustness of the metrics used across the relevance, informativeness and semantic coherence dimensions should also be interpreted with care. Lastly, the experiment was only conducted using the arXiv benchmark dataset as it is the only dataset where all pre-trained weights are publicly available. To encourage similar analysis to be conducted across a wide range of benchmark dataset and model implementation, our evaluation metric toolkit for dataset and model is available at https://github.com/huankoh/long-doc-summarization.
\section{Metrics}
As evaluating generated summary outputs using manual efforts are costly and impractical, the efficient ROUGE metric \cite{lin2004rouge} has long been the standard way of comparing summarization model performance. It measures the lexical overlap between reference and candidate summary and the common n-gram measures are unigram (ROUGE-1), bigram (ROUGE-2) and longest common sub-sequence (ROUGE-L). However, as it is based on exact token matches and overlap between synonymous tokens or phrases will be ignored, the limitation of ROUGE score metrics have been widely explored \cite{kryscinski2019neural,bhandari2020re,chaganty-etal-2018-price,hashimoto-etal-2019-unifying} and many have also attempted to propose more comprehensive content overlap metrics using soft semantic overlap \cite{ganesan2018rouge,zhang2019bertscore}. Further, while content overlap is the fundamental objective of summarization, the quality of a summary, as \citet{gehrmann2018bottom} and \citet{peyrard2019simple} suggested, should be measured in a multi-dimensional way including relevance, factual consistency, conciseness and semantic coherence. Relevance refers to whether the candidate summary contains the main ideas. Factual consistency metric measures whether a candidate summary is factually consistent with the source document. Conciseness measure whether important information is encapsulated in a short and brief manner. Semantic coherence relates to the collective quality and fluency of summary sentences. Based on these quality aspects, the following discusses the research efforts in the wider summarization domain with a focus on the long document summarization research settings at the end of this section.

\subsection{Relevance}\,
\textit{A) Hard Lexical Overlap}\\ As mentioned above, ROUGE score is an efficient way to consider content overlap through hard lexical matching between candidate summary and the ground truth summary. However, as ROUGE only considers exact matching between reference summary and model output, it (a) will penalize models that coin novel wordings and phrases that do not match the wordings in the reference summary, (b) does not consider factual consistency between the model output and the source document and (c) does not directly consider fluency and conciseness of a summary. Finally, ROUGE score also goes against the human approach of clever paraphrasing and summarizing.

\textit{B) Soft Content Overlap}\\ To solve the problem of exact matching of lexical units, \citet{zhang2019bertscore} proposes a model that measures soft overlap between the reference and candidate summary by comparing the contextual BERT embeddings of both summaries. Other variants of this idea include MoverScore \cite{zhao2019moverscore}, Word Mover Similarity and an extension of it, Sentence Mover Similarity \cite{clark2019sentence, kusner2015word}. The soft content overlap metrics often rely substantially on the encoder used to vectorized the candidate and ground truth summary. BERTScore, for example, utilizes BERT as the fundamental pre-trained model to encode its representations. While BERT has been proven to perform amazingly well under many different benchmark settings, its performance under certain domains such as legal or scientific research has not been thoroughly explored. For example, \citet{beltagy2019scibert} fine-tuned BERT on large-scale scientific paper datasets and have found its performance to improve in scientific domains as compared to the BERT-base model. Evidenced by \citet{tejaswin2021well}'s experimental result where BERTScore is found not to discriminate summaries with and without errors well, this questions the use of BERT-base model as the "independent evaluator" of candidate summaries across all domains.

\textit{C) Reference-free Approach}\\
Rather than measuring the quality of candidate summary based on a ground truth summary, reference-free metrics for relevance measure the quality of candidate summary based on pseudo-reference summaries that are generated from source documents. \citet{wu-etal-2020-unsupervised}'s proposed metric requires training samples of high-quality summaries for model supervision while other reference-free metrics can generate metric scores without the use of high-quality summaries as supervisory signals \cite{gao-etal-2020-supert,chen-etal-2021-training}. In section 3 of benchmark datasets, we see that the information covered by a reference summary depends on the data annotation approach as well as the intent of the original authors. We further observe that reference summaries of certain benchmark datasets contain significant noises. If supervised summarization models were to train on datasets with similar issues, they may fit on target summaries that are inconsistent with the expectation and needs of summary readers. A reference-free approach that can bypass the requirement of ground truth summaries would thus be beneficial to the development of models in cases where there is high heterogeneity in summary reader expectation and/or lack of ground truth summary labels. Nevertheless, the use case of reference-free metrics is often limited by the fact that they still require pseudo-reference summaries to be generated by an "independent model". Last but not least, the reference-free approach can also be used to augment the reference-based metrics \cite{hessel-etal-2021-clipscore}.

\subsection{Factual Consistency}
Widespread factual inconsistency in abstractive summarization model outputs greatly limit the potentiality of these abstractive models to be applied in most commercial settings. To this end, automated metrics on factual consistency have been proposed by others \cite{durmus2020feqa, kryscinski-etal-2020-evaluating, maynez2020faithfulness, wang2020asking}, which can be categorized into two different approaches: Entailment Classification and Question Answering.

\textit{A) Entailment Classification Approach}\\ The entailment classification approach evaluates the factual inconsistency of a candidate summary by breaking down the summary into smaller units (e.g., phrases/sentences)
to be verified against the original document. For example, FactCC \cite{kryscinski-etal-2020-evaluating} implements a BERT-based factual consistency classifier that is trained on synthetic data, where the positive data labels are non-paraphrased and paraphrased sentences from the original source document, and the negative labels are artificially corrupted sentences from the source document. At the inference stage, the faithfulness score for a candidate summary is the number of consistent sentences divided by the total number of summary sentences. Similarly, other proposed models implement factual consistency classifiers by incorporating structured knowledge such as OpenIE triples \cite{goodrich2019assessing} or dependency arc \cite{goyal-durrett-2020-evaluating}. For the classifiers to be effective in discriminating the factual consistency of a candidate summary, they often require supervisory signals from factually consistent and inconsistent data \cite{pagnoni-etal-2021-understanding}.

\textit{B) Question-Answering Approach}\\ The Question-Answering (QA) approach employs a question-generation model to generate questions from a given summary output \cite{durmus2020feqa,wang2020asking}. The generated questions are then answered in two different ways: i) answering the question conditioning on the source text and ii) answering the question conditioning on the summary output. If the answers match between the source text and candidate summary, the answer is then considered consistent, otherwise, it is inconsistent. The final score will be based on discrepancies between the answers generated conditional on the candidate summary and the answers generated conditional on the souce document. Recently, QAGen \cite{nan2021improving} proposes to generate questions and answers from a given text concurrently within a single model to evaluate factual consistency to improve the efficiency of this approach.

\textit{Other Important Studies:}\\ 
It is important to note that the aforementioned works consider factual consistency as a binary outcome. In contrast, FRANK \cite{pagnoni-etal-2021-understanding} advocates for a multi-dimensional approach to evaluate factual consistency based on semantic error, discourse error and content verifiability error. Through substantial human annotation, the study further found that the effectiveness of metrics is found to be extremely dependent on the types of architecture measured and the benchmark dataset used. Similarly, human evaluation experiments from previous works have shown conflicting and varying results in the desired approach of developing factuality metrics \cite{maynez2020faithfulness,nan2021improving}. In response, \citet{gabriel-etal-2021-go} proposed five conditions for the development of an effective factuality metric to encourage better standardization in the factual consistency metric research. 

\subsection{Conciseness and Semantic Coherence}
Metrics to measure other aspects of summarization such as conciseness and semantic coherence were also introduced. As they are not as crucial as relevance and factual consistency, these metrics often complement the others to allow a metric or a model to be more holistic and practical. For example, \citet{bommasani2020intrinsic} considers semantic coherence of reference summaries when evaluating single document benchmark datasets while \citet{ju2021leveraging}'s unsupervised model generates fluent summary by utilizing the next sentence prediction task in BERT. Metrics for conciseness are also introduced to measure the quality of summaries \cite{bommasani2020intrinsic,chen-etal-2021-training}.

\subsection{Research Efforts on Metrics in the Long Document Domain}
Many recently proposed metrics incorporate pre-trained architectures to achieve better performances. However, as argued in the model discussion above, these pre-trained architectures cannot be easily extended to long documents. As an illustration, our experiment has attempted various adaptations\footnote{Result details are in the Supplementary Materials.} on a BERT textual entailment model to evaluate arXiv candidate summaries but has found it not effective in discriminating a summary's factual consistency with the source. This is despite \citet{maynez2020faithfulness}'s finding that this model best correlates with the human judgment of factual consistency on the XSUM short document dataset. Furthermore, other than the difficulty of adapting these models on long documents, \citet{nan2021improving} has also identified the issue of resource efficiency, where a competing model would take approximately 4 days to evaluate a CNN-DM test set with an NVIDIA V100 Tensor Core GPU and would likely take significantly longer under any long document benchmark datasets. Consequently, the need of re-designing the proposed evaluation models and the requirement for costly computation resources have likely discouraged the adoption of factual consistency assessment models in the long document summarization domain. Looking at the broader research on evaluation metrics of summarization as a whole, for 17 different research papers related to evaluation metrics published in ACL main conferences\footnote{ACL main conferences are ACL, NAACL, EACL, EMNLP, CoNLL, and AACL. Papers are listed in the Supplementary Materials.} from 2015 to September 2021, there were no discussion on the evaluation metrics in the context of long document summarization datasets. This is important as \citet{pagnoni-etal-2021-understanding} has found that the effectiveness of proposed metrics to vary based on the dataset characteristics. In sum, unlike the quick adoption of short document practices in the model architectures space, research in exploring evaluation metrics within the context of long document summarization is lacking and may potentially hold back the future progression of long document summarization. 
\section{Applications}
As the quality of long document summaries generated by state-of-the-art models continues to improve, past works have explored their feasibility in the research and industrial domains. A natural extension for models that were implemented on the scientific paper benchmark, arXiv/PubMed, is to employ it for research purposes. These include writing section-structured \cite{meng2021bringing}, user-specific \cite{he2020ctrlsum} or presentation-based \cite{sun2021d2s} summaries for scientific papers, automating scientific reviewing \cite{yuan2021can}, and even generating literature survey based on multiple biomedical long scientific papers \cite{deyoung-etal-2021-ms}. When it comes to the general industrial applications of long document summarization, the knowledge and techniques learned from the research domain can address numerous commercial tasks. On the surface level, any information that would be expressed in a textual format would benefit from the advancement in this field, which encompasses summarizing any forms of long textual documents  \cite{sharma2019bigpatent,loukas2021edgarcorpus}, extracting content as feature snippets for search engines\footnote{https://developers.google.com/search/docs/advanced/appearance/featured-snippets}, writing reviews for long media content \cite{kryscinski2021booksum} and summarizing long dialogues \cite{liu2019automatic,chintagunta2021medically,zhong-etal-2021-qmsum,zhu2021mediasum} and multi-modal content \cite{yu2021vision}. With the development becomes increasingly mature in the real-world settings, summarization models are now commercialized as a Software-as-a-Service (Saas) product in the news\footnote{https://ai.baidu.com/tech/nlp\_apply/news\_summary}, business\footnote{https://quillbot.com/} and consulting\footnote{https://www.datagrand.com/about-us/} domains. Furthermore, as the long document summarization task can be generally understood as identification of important aspects from long sequences, the positive spillover from successful model implementation in this domain can affect a wide range of domains. Long document summarization models, for example, can be utilized for auxiliary tasks such as video captioning \cite{liu2021video}, long document question-answering \cite{lyu2021improving} or multi-modal tasks \cite{li2020maec,narasimhan2021clip}.  \citet{liu2018generating} also identified the "unexpected side-effect" of language model reliably learned how to transliterate names between languages, despite the fact that the model was trained to summarize long Wikipedia articles, while BigBird \cite{zaheer2020big} applies Transformer-based models designed for long sequences not only to long document summarization but also to DNA promoter region and chromatin profile prediction tasks in the genomics research domain.
\section{General Challenges and Future Directions}
This section discusses the general challenges of long document summarization that have yet to be solved and pinpoints potential future research directions to attract practitioners' attention and improve our understanding and techniques in the long document summarization domain. Advancement in the long document summarization domain should also give rise to beneficial spillover to closely-related NLP sub-domains such as multi-hop QA, information retrieval and reading comprehension.

\subsection{Neural Models and Long Sequence Reasoning}
While there have been significant efforts in solving the time and memory complexity of a neural architecture such as Transformers to enhance model efficiencies, the understanding of a model's effectiveness in solving different NLP tasks or domains is limited. As shown by our model experiment and result findings of others \cite{zaheer2020big,beltagy2020longformer,huang2021efficient}, fine-tuning pre-trained models using efficient Transformers that can attend to larger input size of tokens can improve the model performances across a wide range of NLP tasks. However, the underlying reasons of the performance improvement is not well understood. For example, while Transformer models have found to outperform RNN models as RNN lacks the ability to reason over long sequences, \citet{pagnoni-etal-2021-understanding} have found that pre-trained Transformer summarization models still make a similar amount of discourse-related errors as the RNN models. Furthermore, research in the effectiveness of various efficient attention mechanisms used by a Transformer to summarize long documents also showed varying results. On the one hand, \citet{huang2021efficient} showed that efficient attention with learnable patterns to significantly outperform the the efficient attention with fixed patterns such as local-only attention mechanism. On the other hand, \citet{manakul2021longspan} have found that extending window size of efficient Transformers to increase number of attended tokens per token do not affect the average distance of attended neighbor, suggesting that local attention to neighboring tokens will be sufficient for the long document summarization task. Altogether, these results highlight the limited understanding on the strategies employed by current neural models to summarize long document and the need of further research to enhance our understanding on this issue.

\subsection{Summarizer with Automatic Discourse Parsers/Annotator}
Our experimental result has demonstrated that the simple unsupervised graph architecture outperforms the other unsupervised models when discourse bias of arXiv section information is included. Nonetheless, information regarding sections of a document may not always be of high quality or available for a summarization model. This limits the implementation of many long doument summarization models that require explicit section-based discourse information. Similar issue has been faced by researchers in the dialogue summarization domain where discourse level information is not provided and past work in this domain have achieved state-of-the-art results by incorporating effective automatic discourse annotators \cite{liu2019automatic,feng-etal-2021-language,wu-etal-2021-controllable}. An architecture that can effectively incorporate automated discourse parsers or annotators would thus be a fruitful direction for long document summarization researchers to explore.

\subsection{End-to-end Neural Summarizer with Content Selection Mechanism}
In the medium term, in spite of the expected progress in computing efficiencies, there exist a significant amount of long documents such as business reports and books that have tokens that exceed hundreds of thousand \cite{loukas2021edgarcorpus,kryscinski2021booksum}. Thus, it is not possible to summarize the entire document using a powerful state-of-the-art model without any long document adaptation, as it will truncates most of the long document source text given the current input length limit. A more practical direction is to explore architectures with a content selection mechanism that has shown to be effective in long document summarization \cite{zhao2020seal,manakul2021longspan}. \citet{zhao2020seal} has proposed an end-to-end long document summarization framework using transformers but did not incorporate powerful pre-trained models and performed slightly worse than other state-of-the-art models. LoBART \cite{manakul2021longspan}, on the other hand, did not design the content selection and abstractive summarizer in an end-to-end manner. Experimental result in this survey has shown that LoBART's disconnection between the retriever and summarizer resulted in less semantically coherent summaries. In the open-domain QA domain, RAG \cite{lewis2020retrieval} achieves state-of-the-art by successfully incorporating content selection mechanism and pre-trained models in an end-to-end manner \cite{petroni2021kilt}, pointing a promising direction for practitioners in the long document summarization domain to explore.

\subsection{Quality and Diversity of Benchmark Dataset}
In section 3 of benchmark datasets, human annotation efforts have been done to measure the quality of the most commonly used long document summarization dataset, arXiv. It was found that 60\% of reference summaries contain some form of errors and 15\% of them have significant errors where at least half of the summary contains errors. This calls for a benchmark dataset with significantly better quality with fewer errors through robust heuristic rules and scraping strategies. Moreover, the long document summarization benchmark datasets are often in the legislative and scientific domain. While these domains are extremely important, many other domains such as financial reports with significant numerical complexity or long-form dialogues of daily conversation in business settings are equally important. Development across different domains could attract even greater attention from a wide range of partners to incentivize greater research efforts in the summarization field. Last but not least, to achieve the original objectives of benchmark datasets, proposed model architectures for long document summarization should also be tested across a diverse set of long document benchmark datasets rather than focusing merely on arXiv/PubMed.  

\subsection{Practicality of Summarization Metrics}
The limitation of ROUGE metric that has been widely explored  \cite{zhou-etal-2006-paraeval,ng2015better,ganesan2018rouge,kryscinski2019neural} and significant efforts have been made to improve the way we measure candidate summaries from various different aspects. Nonetheless, the proposed methods lack practicality in terms of wide availability for all parties in the research communities. For example, \citet{nan2021improving} found that using a single NVIDIA V100 Tensor Core GPU, a factual consistency metric proposed requires longer than four days to evaluate a single set of candidate summaries in the CNN-DM test dataset. Many metrics proposed also require substantial computing resources to re-train across different benchmark settings \cite{kryscinski-etal-2020-evaluating,wang2020asking,durmus2020feqa}. These issues will be exacerbated when it comes to the long document summarization domain. Moreover, most summarization metrics are only tested in the CNN-DM and XSum datasets but not others. This significantly limits its applicability as \citet{pagnoni-etal-2021-understanding} have found most metrics to lack robustness across different benchmark settings. To ensure effective metrics have wider application, efficiencies and practicality of metrics should be paid with great attention to ensure that sufficient incentive is provided for practitioners to explore the practicality of metrics rather than a mere focus on state-of-the-art metric performances. 
\section{Conclusion}
In this survey, we conduct a comprehensive overview of long document summarization and systematically analyze the three key components of its research settings: benchmark datasets, summarization models and evaluation metrics. We first highlight the intrinsic differences of short and long document datasets and show that summarizing long documents requires extra compression of the source text through the identification of key narratives that are more uniformly scattered across the source documents. Nevertheless, long documents are often more extractive in nature and often have explicit discourse structures to take advantage of. For summarization models, we provide a thorough review, comparison and summarization of the model architectures and mechanisms used to generate long document summaries. Through ad-hoc experiment, we also systematically investigate the architectures and mechanisms that are widely applied across various works. We further discuss the current research in evaluation metrics and call attention to the lack of research on metrics that can be easily applied to the long document summarization domain. Finally, we explore the applications of long document summarization models and suggest five future directions for long document summarization research.  
\bibliographystyle{ACM-Reference-Format}
\bibliography{acm-custom}

\newpage
\section{Supplementary materials}

\subsection{Long Document Summarization Systems}
Table \ref{lds_model_summ} details the summary of long document baseline and state-of-the-art summarization systems proposed by previous works in this domain. The "Prior" column illustrates whether inductive bias such as discourse structure information of the original long document is used and the "Trunc" column reflects the total percentage of significant truncation across the five long document benchmark. 

\begin{table*}[h]
\resizebox{\textwidth}{!}{%
\begin{tabular}{c|c|l|c|c}
\hline
\multicolumn{1}{l|}{} &
  \textbf{Model} &
  \multicolumn{1}{c|}{\textbf{Description}} &
  \textbf{Prior} &
  \textbf{Trunc} \\ \hline
\multirow{4}{*}{\textit{\begin{tabular}[c]{@{}c@{}}Unsupervised \\ Baseline \\ Extractive\end{tabular}}} &
  LSA \cite{gong2001generic} &
  Singular Value Decomposition + Term Frequency of Sentence Matrix &
  - &
  - \\
 & TextRank \cite{mihalcea2004textrank}     & Sentence Graph Centrality (PageRank) + Similarity(common words)                    & -       & -    \\
 & LexRank \cite{erkan2004lexrank}       & Sentence Graph Centrality (PageRank) + Similarity(Tf-Idf)                          & -       & -    \\
 & SumBasic \cite{vanderwende2007beyond}      & Average Word-Occurrence Probability of Sentence                                    & -       & -    \\ \hline
\multirow{3}{*}{\textit{\begin{tabular}[c]{@{}c@{}}Unsupervised \\ Neural\\ Extractive\end{tabular}}}
 & PacSum \cite{zheng2019sentence}       & Sentence Graph Centrality + Similarity(BERT)                                       & -       & -    \\
 & HipoRank \cite{dong2021discourse}      & Hierarchical Section-Sentence Graph Centrality + Similarity(BERT)                  & Section & -    \\
 & FAR \cite{liang2021improving} & Sentence Graph Centrality + Similarity(BERT) + Facet Aware Candidate Search        & -       & -    \\ \hline
\multirow{4}{*}{\textit{\begin{tabular}[c]{@{}c@{}}Supervised \\ Neural \\ Extractive\end{tabular}}} &
  Sent-CLF/PTR \cite{pilault2020extractive} &
  Bi-LSTM (word \& sentence level) w/o pretraining &
  - &
  - \\
 & GlobalLocal \cite{xiao2019extractive}   & GloVe (word/sentence level) + Bi-LSTM (section \& document level)                  & Section & -    \\
 & Topic-GraphSum \cite{cui2020enhancing} & BERT (sentence level) + Bipartite GAT (NTM topic-sentence)                         & -       & UNK  \\
 & SSM-DM  \cite{cui2021sliding} & BERT (sentence level) + Sliding Selector Network (Memory Network \& GAT)           & -       & -    \\ \hline
\multirow{6}{*}{\textit{\begin{tabular}[c]{@{}c@{}}Supervised \\ Neural \\ Abstractive\end{tabular}}} &
  Discourse-Aware \cite{cohan2018discourse} &
  BiLSTM (word- \& section- level) w/o pretraining &
  Section &
  - \\
 & Pegasus \cite{zhang2020pegasus}       & Pretraining Summarization Task (GSG) on C4/HugeNews + Data-specific fine-tuning    & -       & 85\% \\
 & CRTLSum \cite{he2020ctrlsum} & Fine-tuned BART with keywords prompt-engineering                                   & -       & 85\% \\
 & BigBird \cite{zaheer2020big}       & Pretrained Pegasus with Sparse Attention (Local + Global + Random)                 & -       & 20\% \\
 & Longformer  \cite{beltagy2020longformer}   & Fine-tuned BART with SparseAttention (Local + Global)                              & -       & -     \\
 & HEPOS \cite{huang2021efficient} & Fine-tuned BART with Efficient Attentions (LSH/Sinkhorn + Hepos)                   & -       & 2\%   \\ \hline
\multirow{4}{*}{\textit{\begin{tabular}[c]{@{}c@{}}Supervised \\ Neural \\ Hybrid \end{tabular}}} &
TLM+Ext \cite{pilault2020extractive}  & Bi-LSTM for ContentSelection + Transformer-based Decoder w/o pretraining           & -       & -    \\ 
& DANCER   \cite{gidiotis2020divide} & Pretrained Pegasus Section-by-Section Summarization                                & Section & -  \\
& SEAL \cite{zhao2020seal}  & End-to-end Transformer-based Content Selector + Decoder w/o pretraining            & -       & -    \\
& LoBART \cite{manakul2021longspan} & Multi-task RNN for ContentSelection + Fine-tuned BART with SparseAttention (Local) & -       & - \\\hline
\end{tabular}%
}
\caption{Summary of Long Document Baseline and State-of-the-Art Summarization Systems.}
\label{lds_model_summ}
\end{table*}

\subsection{Graph-based Ranking Algorithm in Experimental Section}
In general form, given a set of sentences in the original source document, $D = \{s_1, s_2,...,s_m\}$ with the inter-sentential similarity relations represented as $e_{ij} = (s_i,s_j) \in E$  where $i \neq j $, the following equation illustrates the graph-based ranking architecture in computing the scoring for each sentence: 
$$centrality(s_i) = \sum_{j \in \{1,...,m\},i\neq j} e_{ij} * Bias(e_{ij})$$  

The similarity between each sentence is computed using similarity measures such as cosine similarity after being encoded using a sentence encoder. The graph architecture we implemented is a basic directed graph where the centrality score of each sentence is computed based on the summation of bias-adjusted cosine similarity between other sentences and/or sections. The section node will still be represented by sentences where it is the average of the representations for sentences within the section of interest. In other words, a sentence that has the highest sum of similarity against all the other sentences after adjusting for bias will be ranked as the top sentence.

\paragraph{Tf-Idf Only} 
For Tf-Idf encoding, we use scikitlearn Tf-Idf vectorizer to train the encoder using source documents in the arXiv test set (note: this does not include the reference summaries). The preprocessing follows \cite{manakul2021longspan} to ensure consistency and the minimum document frequency is set to be 0.01. As the vector dimension based on original Tf-Idf will be huge, we reduce the dimension to 768 using TruncatedSVD. 

\paragraph{BERT-only} For BERT encoding, we utilize SentenceTransformer package (https://www.sbert.net/) where the model used is "bert-base-nli-mean-tokens". 

\paragraph{Tf-Idf + Long Document Discourse Bias} 
Following the implementation of \cite{dong2021discourse}, the bias is calculated using intra-section bias (or position-level bias within each section) and inter-section bias (or section-level bias). For sentences within the same section, the cosine-similarity between sentences adjusted by intra-section bias, $e_{ij}^{intraB}$, is computed by adjusting the lambda biases, $\lambda_1$ and $\lambda_2$ based on sentence boundary function, $d_b$: 
\begin{equation*}
e_{ij}^{intraB} = \left\{
        \begin{array}{ll}
            sim(s^I_j,s^I_i) * \lambda_1, if \,\, d_b(s^I_i) \geq d_b(s^I_j) \\
            sim(s^I_j,s^I_i) * \lambda_2, if  \,\, d_b(s^I_i) < d_b(s^I_j) 
        \end{array}
    \right.
\end{equation*}
where the sentence boundary function, $d_b$, will determine sentences $s^I_i$ that are closer to the section $I$ boundaries to be more important. This is computed by:
$$ d_b(s^I_i) = min(x^I_i,\alpha(n^I - x_i^I))$$
$n^I$ is the number of sentences in section $I$ and $ x_i^I $ represents sentence i’s position in section I. Cosine similarity between sentence and section adjusted by inter-section bias is calculated similarly except that the bias is computed based on the section’s position in the document. Finally, the resulting adjusted centrality score for each sentence is: 
$$c(s_i^I) = \mu_1 \cdot c_{inter}(s_i^I) + c_{intra}(s_i^I)$$
where $\mu_1$ is a weighting factor for inter-section centrality.
\\
\\
For hyperparameter tuning, we set $\lambda_1 = 0.5$ and $\lambda_2 = 1$. Then, using arXiv validation samples, we adjust $ \alpha \in \{0, 0.5, 0.8, 1.0, 1.2\}$ to control the relative importance of the start and end of a section
of a source document and $\mu_1 \in \{0.5, 1.0, 1.5\}$ to control the weights of intra-section sentence importance versus inter-section sectional importance. Importantly, the original paper uses $\lambda_1 = 0 $ where less important sentences are pruned, while we set $\lambda_1 = 0.5$ to down weight rather than prune the less important sentences. For more details, we refer our reader to the original paper \cite{dong2021discourse} with their codes available on https://github.com/mirandrom/HipoRank. 

\paragraph{BERT + Long Document Discourse Bias} Same as Tf-Idf except that the Tf-Idf sentence encoder is replaced by BERT.
\\\\
Lastly, for the implementation of Transformer-based abstractive summarization models, the links to original author's pre-trained weights, codes and implementations are provided in the footnote of our main article.

\subsection{BERT NSP - Assessing Semantic Coherence of Candidate Summaries} \citet{bommasani2020intrinsic} suggested using the general pre-trained BERT model to evaluate the semantic coherence or fluency of a summary without any fine-tuning. We find the general pre-trained BERT model that was not trained on academic papers to have little discriminative ability for the semantic coherence of arXiv-related summaries. Thus, we fine-tune the pre-trained BERT model using positive and negative sentence pairs. Positive sentences are extracted by taking any sentence together with its following sentence in the reference summary and source text of the arXiv dataset. Negative sentences are created by replacing the following sentences with any randomly extracted sentences from either the same or other documents in the arXiv benchmark dataset. We also ensured that the total samples of positive and negative sentence pairs are balanced and an equal amount of sentences are obtained from the reference summary and the source text. The codes of all the metrics used in the main paper, including BERT NSP and intrinsic characteristics of the dataset, are made publicly available on: https://github.com/huankoh/long-doc-summarization.

\subsection{Textual Entailment as Factual Consistency Metric}

\begin{table}[h!]
\small
\resizebox{0.55\textwidth}{!}{%
\begin{tabular}{c|c|c}
\hline
\textbf{Adaptation}                                   & \textbf{Good} & \textbf{Bad} \\ \hline
Ordered R1   &    0.03 /0.10 /0.86                                          & 0.04 / 0.11 / 0.85 \\
Ordered R2 & 0.05 / 0.11 / 0.84 & 0.04 / 0.13 / 0.83 \\
Ordered RL  & 0.04 / 0.12 / 0.84 & 0.04 / 0.13 / 0.83\\
Randomised R1 & 0.04 / 0.25 / 0.71 & 0.04 / 0.30 / 0.66 \\
Randomised R2 & 0.06 / 0.21 / 0.73 & 0.07 / 0.19 / 0.74 \\
Randomised RL & 0.05 / 0.25 / 0.70  & 0.05 / 0.24 / 0.71\\ \hline
\end{tabular}%
}
\caption{Textual Entailment Result (Entail/Contradict/Neutral) on annotated arXiv reference summaries after Long Document Adaptation. The Good column represents the results of reference summaries that are annotated to be high quality and the Bad column for low-quality data.}
\label{tab. fc_table}
\end{table}

Conditional on the source text, the entailment task classifies summary sentences as entails, neutral or contradicts. Ideally, a candidate summary should entail or be neutral to the source text, but never contradict the source text. As BERT textual entailment model without long document adaptation have a token limit of 512, this will not be directly applicable for long document datasets with a token length of at least in the thousands. To use this in our experiment, we attempted to adapt \citet{maynez2020faithfulness}'s textual entailment BERT model by implementing a content selection mechanism to reduce the input size of the source document. The selected subset is constructed using gold label sequences by greedily optimizing the ROUGE score on the ground truth reference summaries, following the algorithm provided by \citet{xiao2019extractive}. As we have different variants of the ROUGE score, we experimented with a greedy selection of Rouge-1, Rouge-2 and Rouge-L in the originally ordered sentences and in the randomized ordered sentences. The aim of randomized ordered sentences is to ensure that the salient contents are extracted more uniformly from the source text. To evaluate the discriminative ability of our adapted model, we use the annotated test data in section 3 to evaluate the discriminative ability of our adapted BERT textual entailment model. As the annotated data has 15\% of the randomly sampled data to be extremely low quality and 30\% to have zero errors in the sentences. A good BERT textual entailment model should then evaluate the 30\% high-quality samples as low contradiction and high entailment or neutrality while evaluating the 15\% low-quality samples with higher contraction and lower entailment or neutrality. From Table \ref{tab. fc_table}, despite trying out various adaptations, our models have almost no discriminative ability and were thus not used in our experimental section in the main article. This is an important limitation of our experiment in section 5 of our main article.

\subsection{Metric-related ACL main conference research papers} Table \ref{tab. metric_paper} next page details the 17 papers published from 2015 to September 2021 in ACL main conferences, including ACL, NAACL, EACL, EMNLP, CoNLL, and AACL.
\begin{table*}[!ht]
\small
\resizebox{0.96\textwidth}{!}{%
\begin{tabular}{c|c}
\hline
\textbf{Paper Title}                                   & \multicolumn{1}{c}{\textbf{Year}} \\ 
                                                 & \\ \hline
\multicolumn{1}{l|}{\textit{I. ACL}}      &                                         \\
Studying Summarization Evaluation Metrics  in the Appropriate Scoring Range & 2019 \\
Facet-Aware Evaluation for Extractive Summarization                                      & 2020                                     \\
FEQA: A Question Answering Evaluation Framework for Faithfulness Assessment in Abstractive Summarization  & 2020 \\
On Faithfulness and Factuality in Abstractive Summarization & 2020
\\ 
Improving Factual Consistency of Abstractive Summarization via Question Answering & 2021\\
A Training-free and Reference-free Summarization Evaluation Metric via Centrality-weighted Relevance and Self-referenced Redundancy & 2021\\
GO FIGURE: A Meta Evaluation of Factuality in Summarization & 2021\\
Focus Attention: Promoting Faithfulness and Diversity in Summarization & 2021 \\
Evaluating the Efficacy of Summarization Evaluation across Languages & 2021 \\
\multicolumn{1}{l|}{\textit{II. NAACL}}             &                                         \\
Question Answering as an Automatic Evaluation Metric for News Article Summarization &  2019                                      \\
Understanding Factuality in Abstractive Summarization with FRANK: A Benchmark for Factuality Metrics          &  2021                                     \\ 
\multicolumn{1}{l|}{\textit{III. EMNLP}} &                                         \\
Re-evaluating Automatic Summarization with BLEU and
192 Shades of ROUGE & 2015 \\ 
Better Summarization Evaluation with Word Embeddings for ROUGE & 2015 \\
Answers Unite! Unsupervised Metrics for Reinforced Summarization Models & 2019 \\
What Have We Achieved on Text Summarization? &  2020                                       \\
Evaluating the Factual Consistency of Abstractive Text Summarization &  2020                                       \\ 
Re-evaluating Evaluation in Text Summarization & 2020\\
\hline
\end{tabular}%
}
\caption{List of Metric-related ACL Main Conferences Papers. EACL, CoNLL, and AACL do not have metric-related summarization research papers.}
\label{tab. metric_paper}
\end{table*}

\end{document}